\documentclass[nohyperref]{article}

\usepackage{microtype}
\usepackage{graphicx}
\usepackage{subfigure}
\usepackage{booktabs} 

\usepackage{hyperref}
\usepackage{amsbsy}

\usepackage[accepted]{icml2022}

\usepackage{amsmath}
\usepackage{amssymb}
\usepackage{mathtools}
\usepackage{amsthm}

\usepackage[capitalize,noabbrev]{cleveref}

\usepackage{wrapfig}
\usepackage{stmaryrd}
\usepackage{enumerate}

\usepackage{amsmath,amsfonts,bm}
\usepackage{multirow}
\usepackage{tabularx}
\usepackage{graphicx}
\usepackage{adjustbox}
\usepackage{amsthm}

\newcommand{\cut}[1]{}

\newcommand{\mset}[1]{\left\{\kern-.5em\left\{ #1 \right\}\kern-.5em\right\}}
\newcommand{\mmset}[1]{\{\kern-.4em\{ #1 \}\kern-.4em\}}

\newcommand{\bb}{\,\|\,}

\newcommand{\data}{\mathrm{data}}

\newcommand{\norm}[1]{\left\Vert#1\right\Vert}

\newcommand{\abs}[1]{\left\vert#1\right\vert}

\newcommand{\set}[1]{\left\{#1\right\}}
\newcommand{\parr}[1]{\left (#1\right )}
\newcommand{\brac}[1]{\left [#1\right ]}
\newcommand{\ip}[1]{\left \langle #1 \right \rangle }
\newcommand{\Real}{\mathbb R}

\newcommand{\too}{\rightarrow}

\newcommand{\eg}{{e.g.}}
\newcommand{\ie}{{i.e.}}

 \newtheorem{theorem}{Theorem}
 \newtheorem{lemma}{Lemma}

 \newtheorem{definition}{Definition}

\def\eqref#1{equation~\ref{#1}}

\def\1{\bm{1}}

\def\vec1{{\bm{1}}}

\def\mI{{\bm{I}}}

\DeclareMathAlphabet{\mathsfit}{\encodingdefault}{\sfdefault}{m}{sl}
\SetMathAlphabet{\mathsfit}{bold}{\encodingdefault}{\sfdefault}{bx}{n}

\def\gM{{\mathcal{M}}}
\def\gN{{\mathcal{N}}}

\def\gS{{\mathcal{S}}}

\def\gU{{\mathcal{U}}}

\newcommand{\dist}{\mathrm{d}} 

\newcommand{\E}{\mathbb{E}}

\DeclareMathOperator{\sign}{sign}

\theoremstyle{plain}

\theoremstyle{remark}

\usepackage[textsize=tiny]{todonotes}

\icmltitlerunning{Matching Normalizing Flows and Probability Paths on Manifolds}

\begin{document}

\twocolumn[
\icmltitle{Matching Normalizing Flows and Probability Paths on Manifolds}



\icmlsetsymbol{equal}{*}

\begin{icmlauthorlist}
\icmlauthor{Heli Ben-Hamu}{equal,wis}
\icmlauthor{Samuel Cohen}{equal,meta,ucl}
\icmlauthor{Avishek Joey Bose}{meta}
\icmlauthor{Brandon Amos}{meta}
\icmlauthor{Aditya Grover}{meta} \\
\icmlauthor{Maximilian Nickel}{meta}
\icmlauthor{Ricky T. Q. Chen}{meta}
\icmlauthor{Yaron Lipman}{wis,meta}
\end{icmlauthorlist}

\icmlaffiliation{wis}{Weizmann Institute of Science}
\icmlaffiliation{ucl}{Centre for Artificial Intelligence, University College London}
\icmlaffiliation{meta}{Meta AI Research}

\icmlcorrespondingauthor{Heli Ben-Hamu}{heli.benhamu@weizmann.ac.il}

\icmlkeywords{Machine Learning, ICML}

\vskip 0.3in
]



\printAffiliationsAndNotice{\icmlEqualContribution} 

\begin{abstract}
Continuous Normalizing Flows (CNFs) are a class of generative models that transform a prior distribution to a model distribution by solving an ordinary differential equation (ODE). 
We propose to train CNFs on manifolds by minimizing \textit{probability path divergence} (PPD), a novel family of divergences between the probability density path generated by the CNF and a target probability density path. 
PPD is formulated using a logarithmic mass conservation formula which is a linear first order partial differential equation relating the log target probabilities and the CNF's defining vector field. 
PPD has several key benefits over existing methods: it sidesteps the need to solve an ODE per iteration, readily applies to manifold data, scales to high dimensions, and is compatible with a large family of target paths interpolating pure noise and data in finite time. Theoretically, PPD is shown to bound classical probability divergences.
Empirically, we show that CNFs learned by minimizing PPD achieve state-of-the-art results in likelihoods and sample quality on existing low-dimensional manifold benchmarks, and is the first example of a generative model to scale to moderately high dimensional manifolds.

\end{abstract}

\section{Introduction}

One of the core domains of machine learning research are density estimation and generative modeling, which view data from a probabilistic perspective. The deep-learning revolution fostered a significant advancement in the field, leading to the emergence of powerful generative models for images, language, audio, and other data types represented in Euclidean spaces. While early literature primarily focused on Euclidean data, the need to model data in non-Euclidean spaces arises in many scientific fields. For instance, occurrences of natural phenomena on earth can be modeled as a distribution on a sphere \cite{mathieu2020riemannian}, protein structure prediction requires angle predictions \cite{kanti2008biology} and motion and position of robots can be modeled with a product of Euclidean spaces and spheres. Therefore, constructing generative models over manifolds is an important problem with many potential applications.  

A generative model can be described as a function $\phi$ that transforms a simple probability distribution (the prior or base) to a more complicated one (the model) so to best represent some empirical set of data observations. Among the large toolkit of deep generative models, innate candidates for designing generative models on manifolds are Normalizing Flows (NFs) \cite{Rezende2015nf} and Continuous Normalizing Flows (CNFs) \cite{chen2018neural}. In these approaches $\phi$ is a diffeomorphism, \ie, a smooth bijection with a smooth inverse. Therefore, the model density can be expressed in terms of the prior density and the determinant of the Jacobian of $\phi$, also known as the change of variable formula, which can be naturally adapted to the manifold case. Recently, \cite{rezende2020normalizing, bose2020Latent} devised NF models for sphere, tori and hyperbolic spaces. In a parallel line of works, \cite{mathieu2020riemannian, lou2020neural, falorsi2020Neural} developed CNFs over Riemannian manifolds. 

Like Euclidean NF models, manifold NFs suffer from limited expressive power due to the representation of $\phi$ as a composition of a restricted set of invertible transformations. On the other hand, CNFs model diffeomorphisms as flows along parametric tangent vector fields, $v$, represented as neural networks, lifting the architectural restriction and allowing maximal expressive power. Nonetheless, training a CNF by minimizing negative log-likelihoods, or equivalently, the KL-divergence of the data and model densities, requires log model densities evaluated at observation points. Computing the log model densities entails solving an ordinary differential equation (ODE) during training, which results in a substantial time and memory burden, as well as introduces an extra challenge when the data lies on a manifold. \citet{rozen2021moser} suggested a different parametrization of CNFs via the divergence of unrestricted vector fields, where both training and computing model probabilities do not require solving an ODE. However, scaling this method to even moderately high dimensions is challenging since it is formulated with a density function rather than log density, which can cause numerical issues as density values decrease exponentially with dimension. 


%

This work aims to alleviate some of the limitations of previous approaches by introducing the Probability Path Divergences (PPD),
a new type of divergence defined between an arbitrary target probability path, $p$, and the probability path generated by the CNF, $q$. To define the PPD we first introduce the Logarithmic Mass Conservation (LMC) formula, a Partial Differential Equation (PDE) that couples $\log q$ and the CNF's vector field. Then, the PPD is defined as the extent to which $\log p$ and the CNF's vector field fail to satisfy the LMC. 
PPD has the following desirable properties: (i) It is a proper divergence in the sense that it is non-negative, and zero iff $p \equiv q$. (ii) It does not require evaluating $q$ during training; it is defined solely in terms of the parametric vector field $v$, its first order derivatives, and the target path's log density, $\log p$. This provides a speed up of $1-2$ orders of magnitude in evaluating the PPD and its derivatives, compared to, \eg, log likelihood. (iii) It is readily applicable to manifolds and higher dimensional data. (iv) The PPD has a single parameter $\ell\geq 1$. PPD with $\ell=1$ upper bounds the total-variation divergence comparing $p$ and $q$ at arbitrary times; PPD with $1<\ell<\infty$ bounds their $\alpha$-divergence; and PPD with $\ell=\infty$ bounds their reversed KL-divergence. 

We call the minimization problem of the PPD between a target path $p$ and a CNF density $q$ CNF Matching (CNFM), and use it to train CNFs. The main design choice in CNFM is the target path $p$. The requirements from $p$ are: that it transforms a simple prior (pure noise) to an approximation of the unknown data distribution; that samples can be drawn from each $p_t$, where $p_t$ represents the density at time $t$; and that we can compute or approximate the derivatives of $\log p_t$. Any $p$ satisfying these requirements can be used to train a CNF in the CNF Matching framework. Other methods that try to fit generated probability density path to a target one are Score and Diffusion based methods \cite{song2019generative,ho2020denoising,song2020score}. However, these methods require target paths that are generated by Stochatsic Differential Equations (SDEs) or known diffusion processes which limits their applicability on manifolds. We elaborate this discussion in Section \ref{ss:previous_works_sde}, after introducing our method.

We test our framework on several low and moderately high dimensional manifold data including Euclidean spaces, spheres/hyperspheres, and product of spheres, demonstrating state-of-the-art sample quality and likelihoods in standard low-dimensional manifold datasets. We demonstrate that CNFM is considerably faster to optimize than state of the art CNF training algorithm, allowing to scale CNF training to considerably larger network architectures. Lastly, we demonstrate  that CNFM can train CNFs on moderately high dimensional manifolds, in contrast to previous methods of generative modeling on manifold that mostly worked with low dimensional manifolds. 


    
    
    
    
    

\section{Preliminaries}
Let $\gM$ be a $d$-dimensional smooth Riemannian manifold with a metric $g$ and induced volume form $dV$, the volume of $\gM$ is $\abs{\gM}=\int_\gM dV_x$. We consider strictly positive, smooth probability densities over $\gM$, $\mu:\gM\too \Real_{>0}$, satisfying $\int_\gM \mu(x) dV_x = 1$. 
The tangent space at point $x\in \gM$ is denoted $T_x\gM$; the tangent bundle, which is the disjoint union of all tangent spaces of $\gM$ is denoted $T\gM$. 
The metric $g$ defines an inner product for pairs of vectors $\xi,\eta\in T_x\gM$ denoted by $\ip{\xi,\eta}$; a norm of a tangent vector is defined by $\abs{\xi}=\ip{\xi,\xi}^{1/2}$. The Riemannian gradient of a smooth function $f:\gM\too \Real$ is denoted $\nabla f(x)\in T_x\gM$.  
A time-dependent vector field $v(t,x)$ is a smooth function $v:[0,1]\times \gM\too T\gM$ such that $v(t,x)\in T_x\gM$ for all $t\in[0,1]$ and $x\in \gM$. We denote the collection of bounded time dependent smooth vector fields over $\gM$ by $\mathfrak{X}(\gM)$; by bounded we mean that for each $v\in \mathfrak{X}(\gM)$ there exists a constant $M>0$ so that $\abs{v(t,x)}\leq M$ for all $x\in\gM$, $t\in[0,1]$. The Riemannian divergence (w.r.t.~$x$) of a smooth vector field $v\in \mathfrak{X}(\gM)$ is denoted $\mathrm{div}(v)$.
We denote by $\exp_x:T_x\gM\too \gM$, and $\log_x:\gM\too T_x\gM$ the Riemannian exponential and logarithmic maps. Note these should not be confused with the standard $\exp,\log$ that are written without subscript. 

Given a time dependent vector field $v\in\mathfrak{X}(\gM)$, a one parameter diffeomorphism family $\phi_{t}:\gM\too\gM$ can be defined via the Ordinary Differential Equation (ODE):
\begin{align}\label{e:phi}
\begin{cases}
\frac{d}{dt}\phi_{t}(x) = v(t,\phi_{t}(x)) & \\
\phi_{0}(x) = x & 
\end{cases}
\end{align}
In the context of generative models, the diffeomorphism $\phi_t$ is called a Continuous Normalizing Flow (CNF) \cite{chen2018neural,mathieu2020riemannian,lou2020neural,falorsi2020Neural,rozen2021moser} and is used to push-forward or pull-back probability densities. An event $A\subset \gM$ is pushed forward by $\phi_t$ to the event $\phi_t(A)$, and pulled back to $\phi_t^{-1}(A)$. Given a probability density $\eta$ over $\gM$ its pushed forward density is denoted $\phi_{t*}\eta$, and its pulled back density by $\phi_t$ is denoted $\phi_t^* \eta$. Let $\mathfrak{P}(\gM)$ denote all \emph{probability paths} on $\gM$, that is functions $p:[0,1]\times \gM\too\Real_{> 0}$, smooth in $t$ and satisfying $\int_\gM p(t,x)dV_x=1$.

\begin{definition}\label{def:phi_t_generates_alpha_t}
We say that a CNF $\phi_t$ \emph{generates} a probability density path $q\in \mathfrak{P}(\gM)$ if for all $t\in [0,1]$
\begin{equation}
    q_t=\phi_{t*}q_0, \text{ or equivalently } \phi_t^*q_t=q_0,
\end{equation}
\end{definition}

\begin{figure*}[t]
    \centering
    \vspace{10pt}
    \begin{tabular}{c@{\hspace{1pt}}c@{\hspace{1pt}}c@{\hspace{1pt}}c@{\hspace{1pt}}c@{\hspace{1pt}}c@{\hspace{1pt}}c@{\hspace{1pt}}c@{\hspace{1pt}}c@{\hspace{1pt}}c}
        \includegraphics[width=0.14\textwidth]{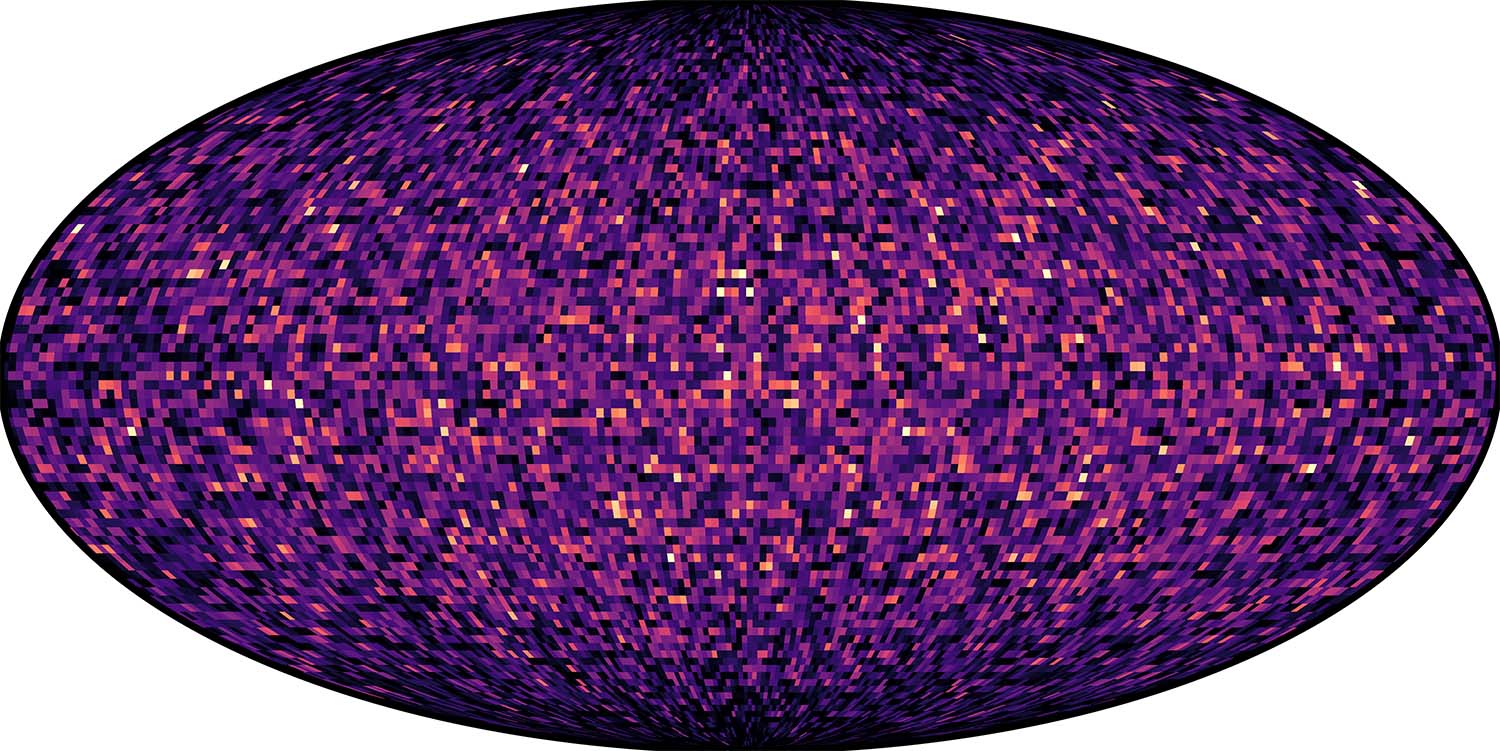} & 
        \includegraphics[width=0.14\textwidth]{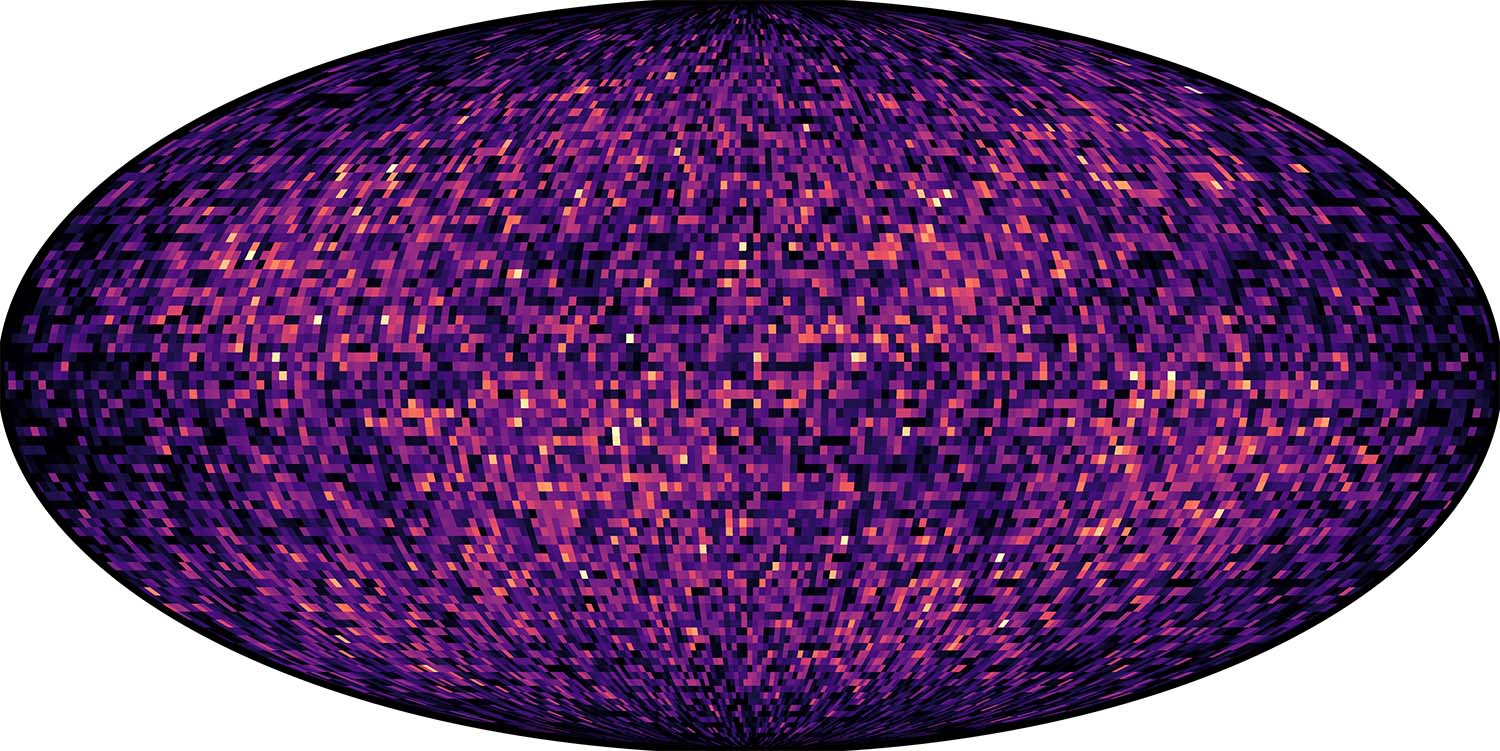} & 
        \includegraphics[width=0.14\textwidth]{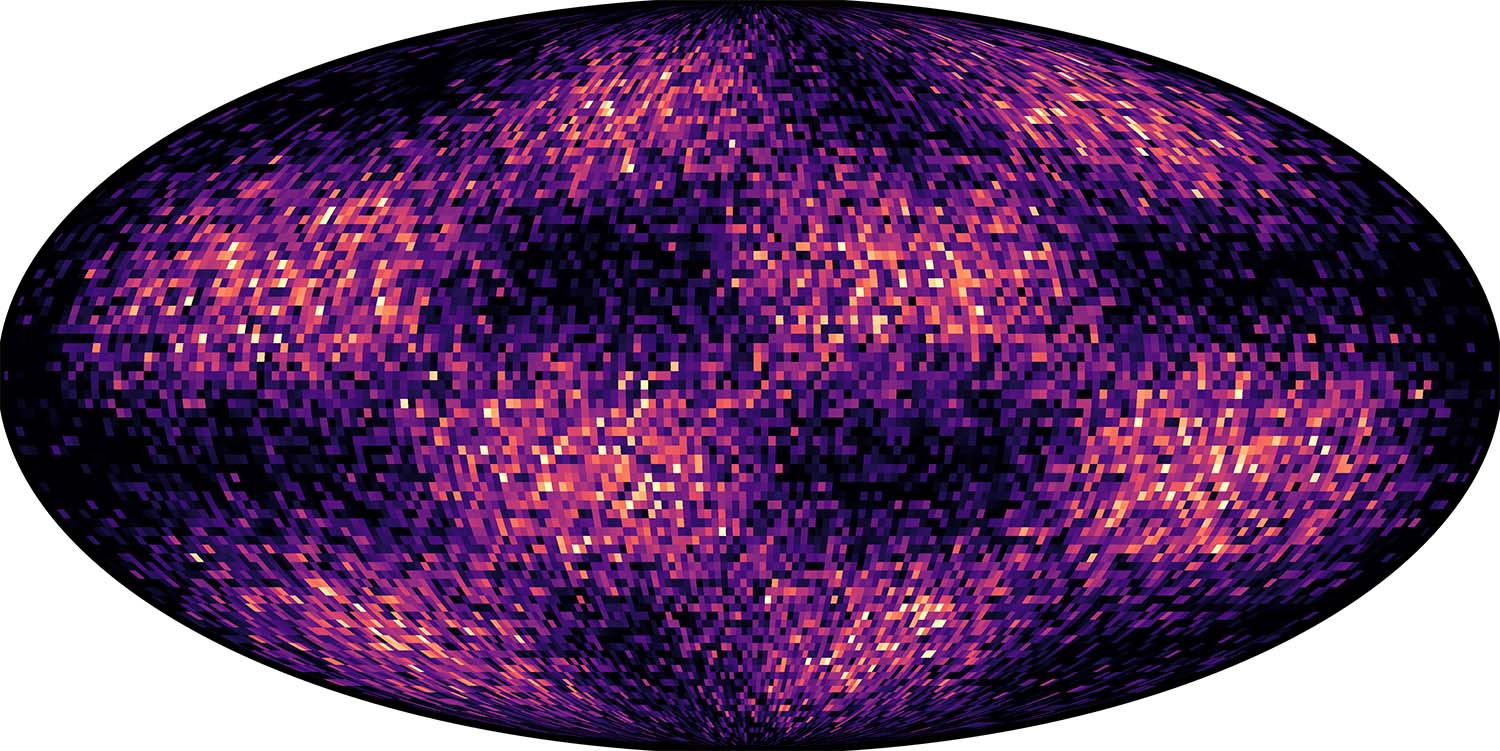} & 
        \includegraphics[width=0.14\textwidth]{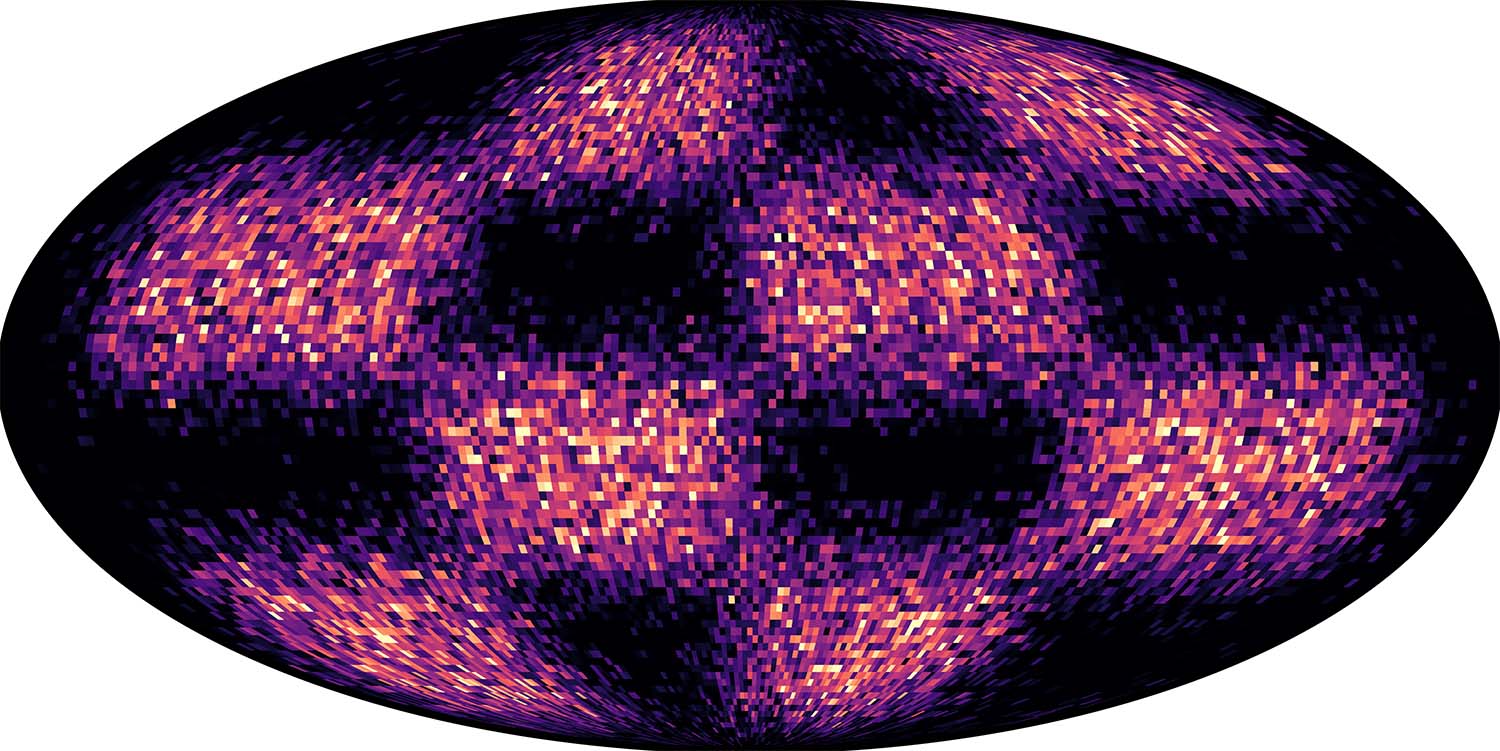} & 
        \includegraphics[width=0.14\textwidth]{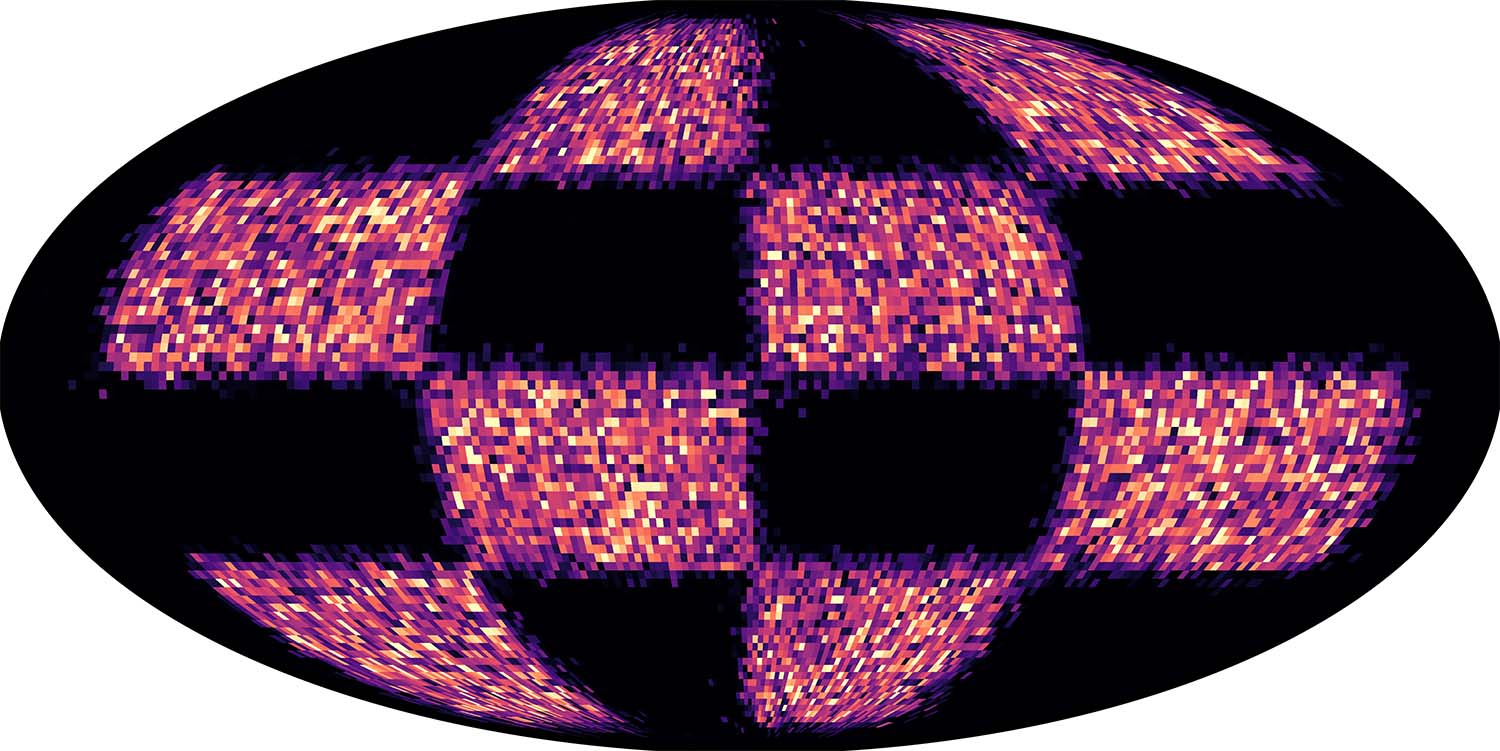} & 
        \includegraphics[width=0.14\textwidth]{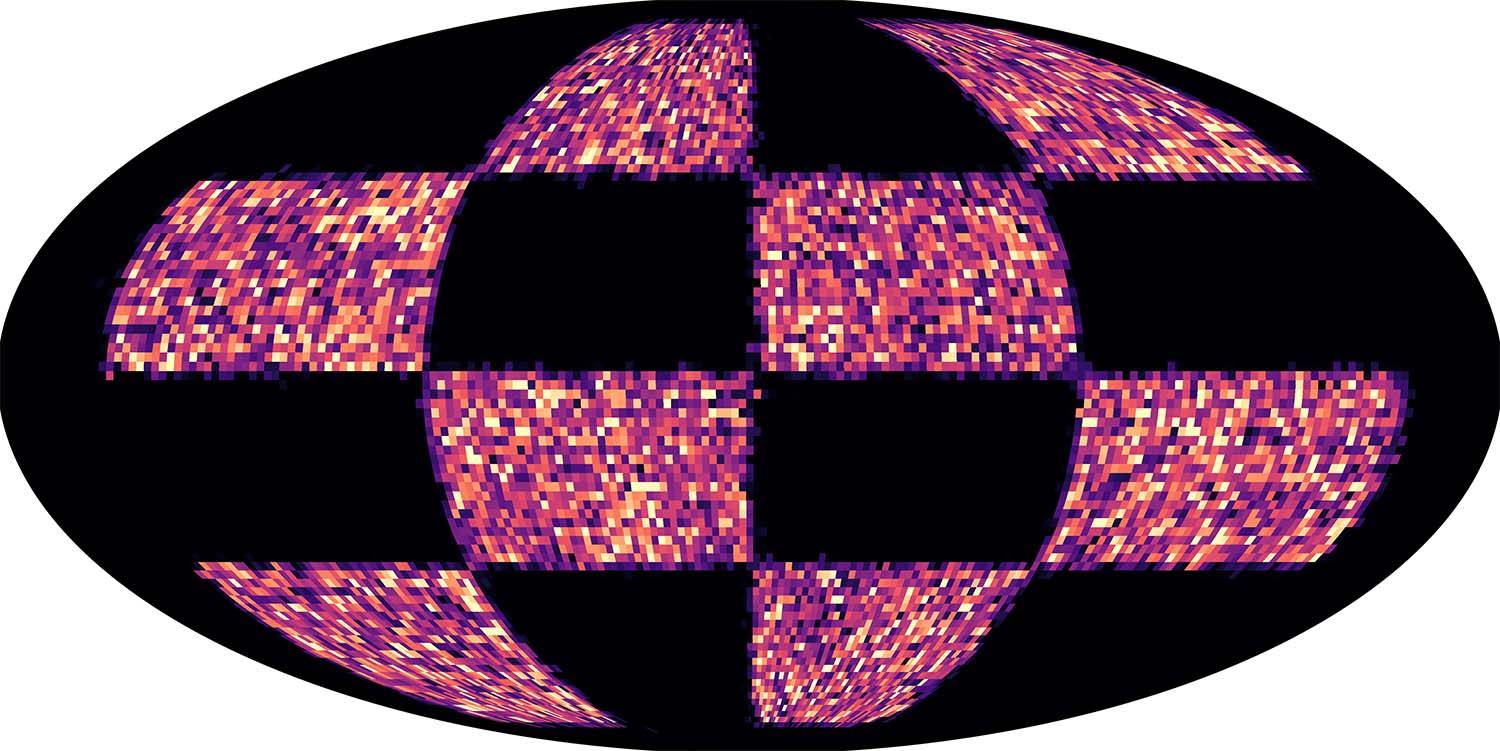} & 
        \includegraphics[width=0.14\textwidth]{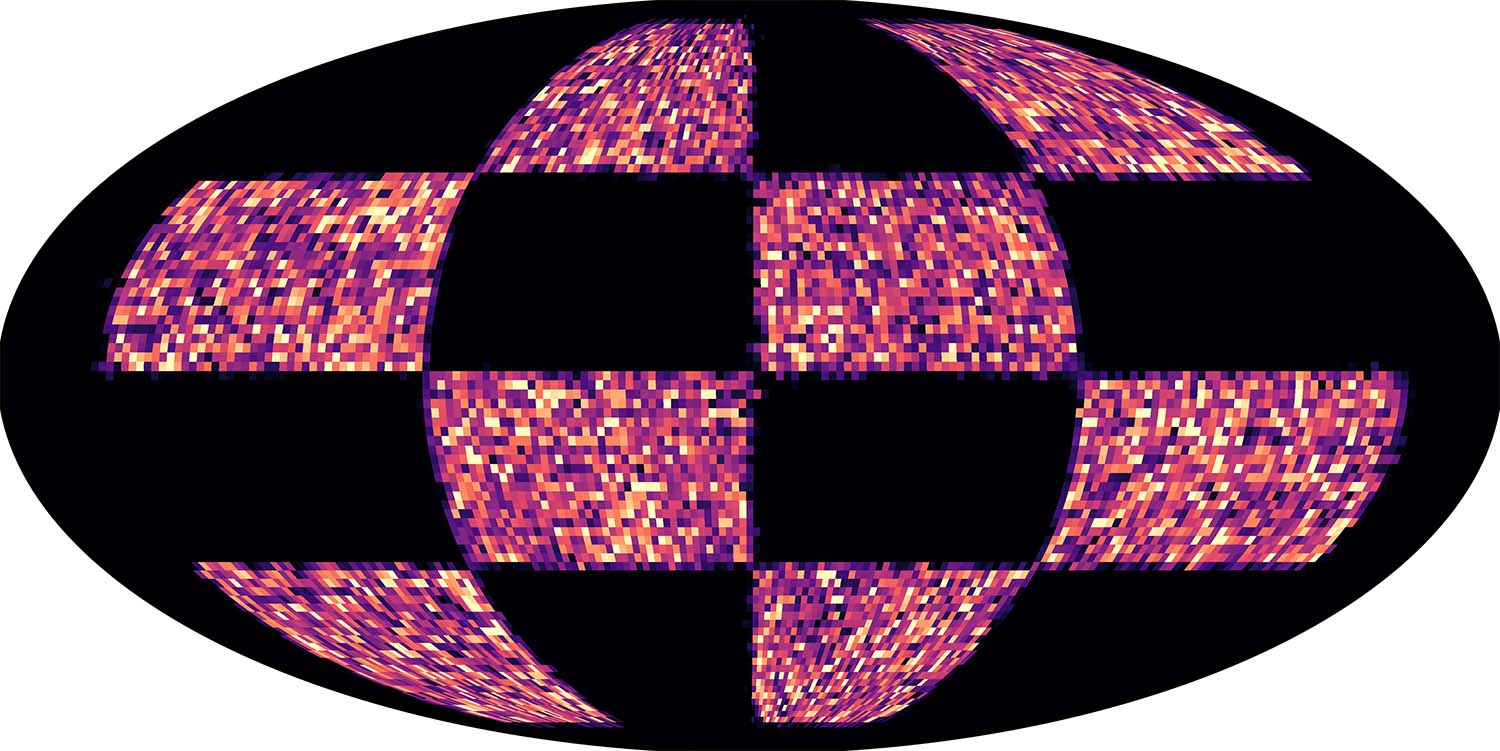} \\
        \includegraphics[width=0.14\textwidth]{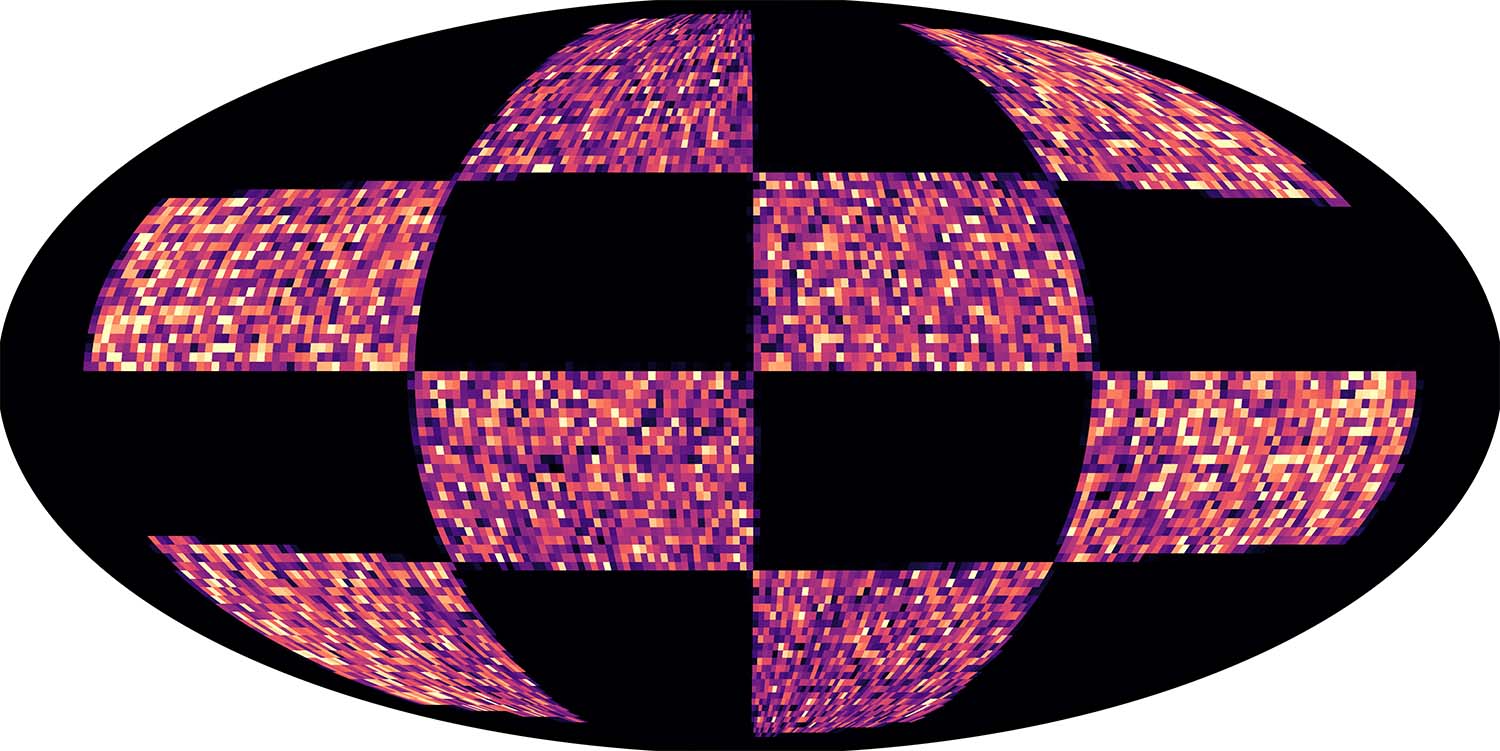} & 
        \includegraphics[width=0.14\textwidth]{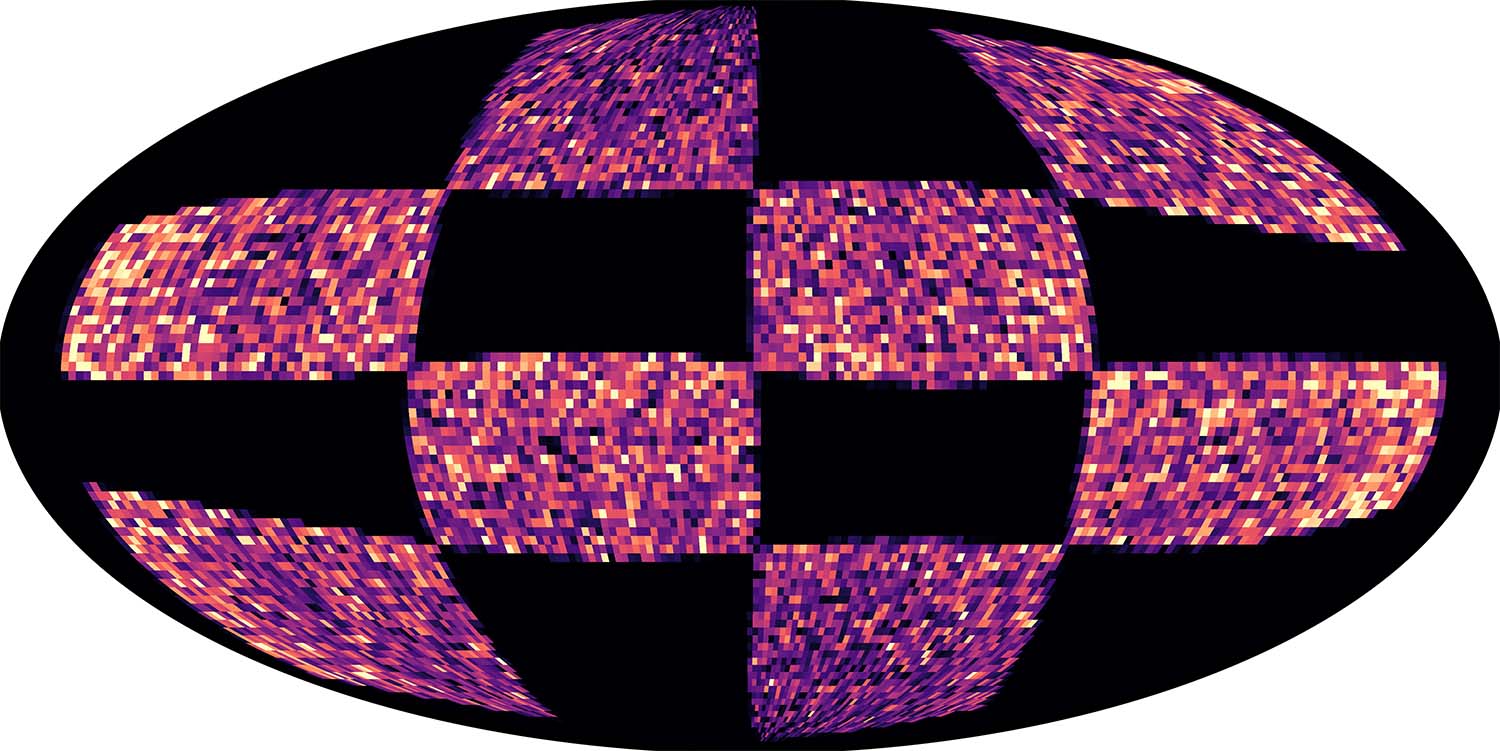} & 
        \includegraphics[width=0.14\textwidth]{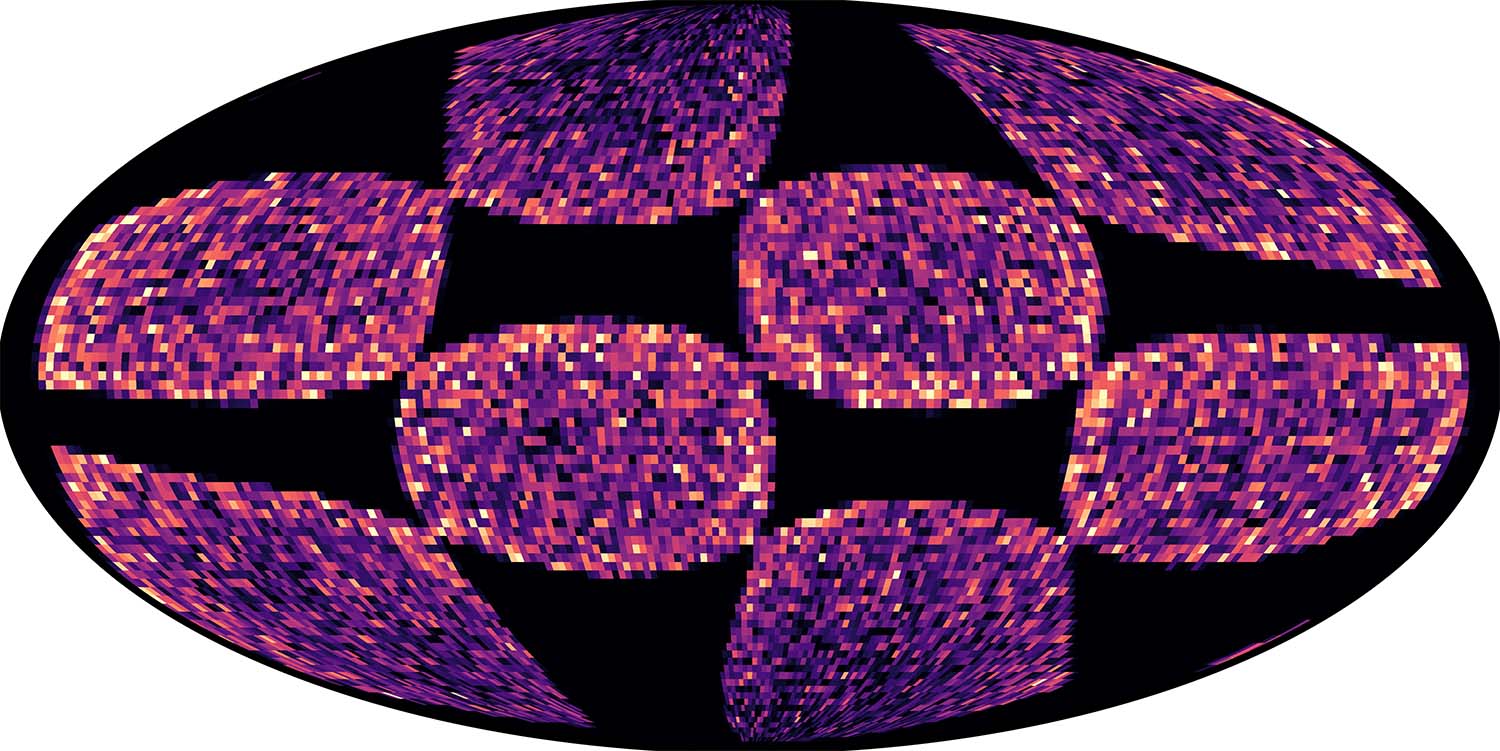} & 
        \includegraphics[width=0.14\textwidth]{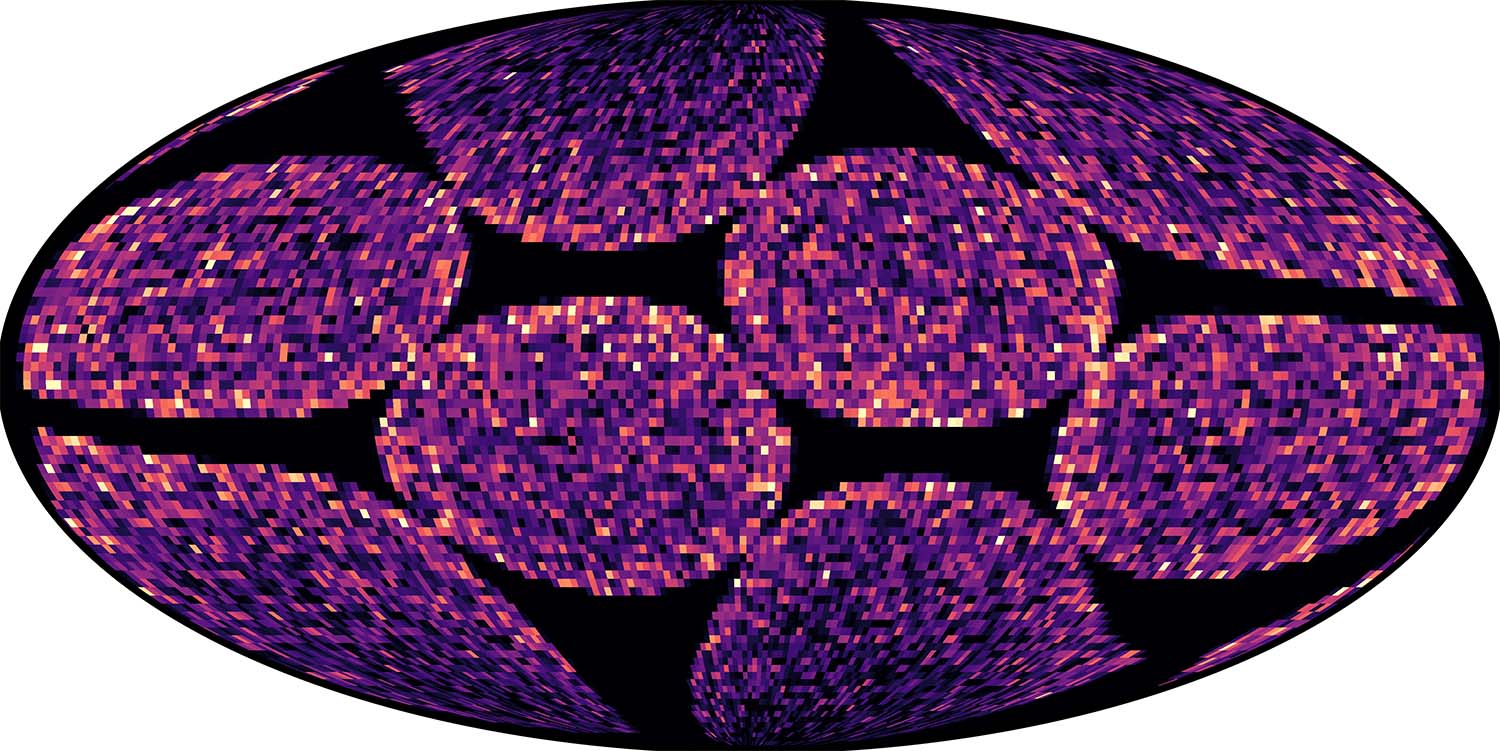} & 
        \includegraphics[width=0.14\textwidth]{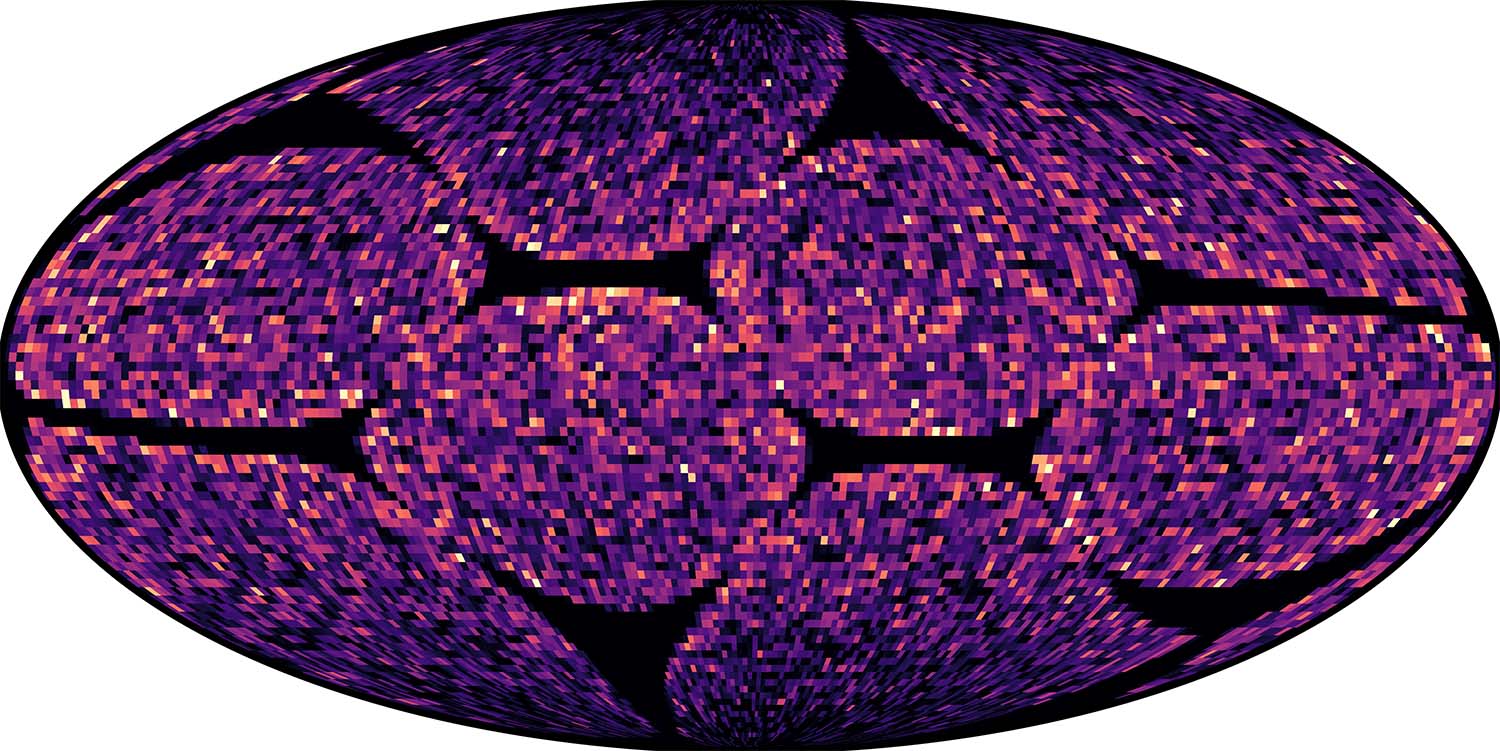} & 
        \includegraphics[width=0.14\textwidth]{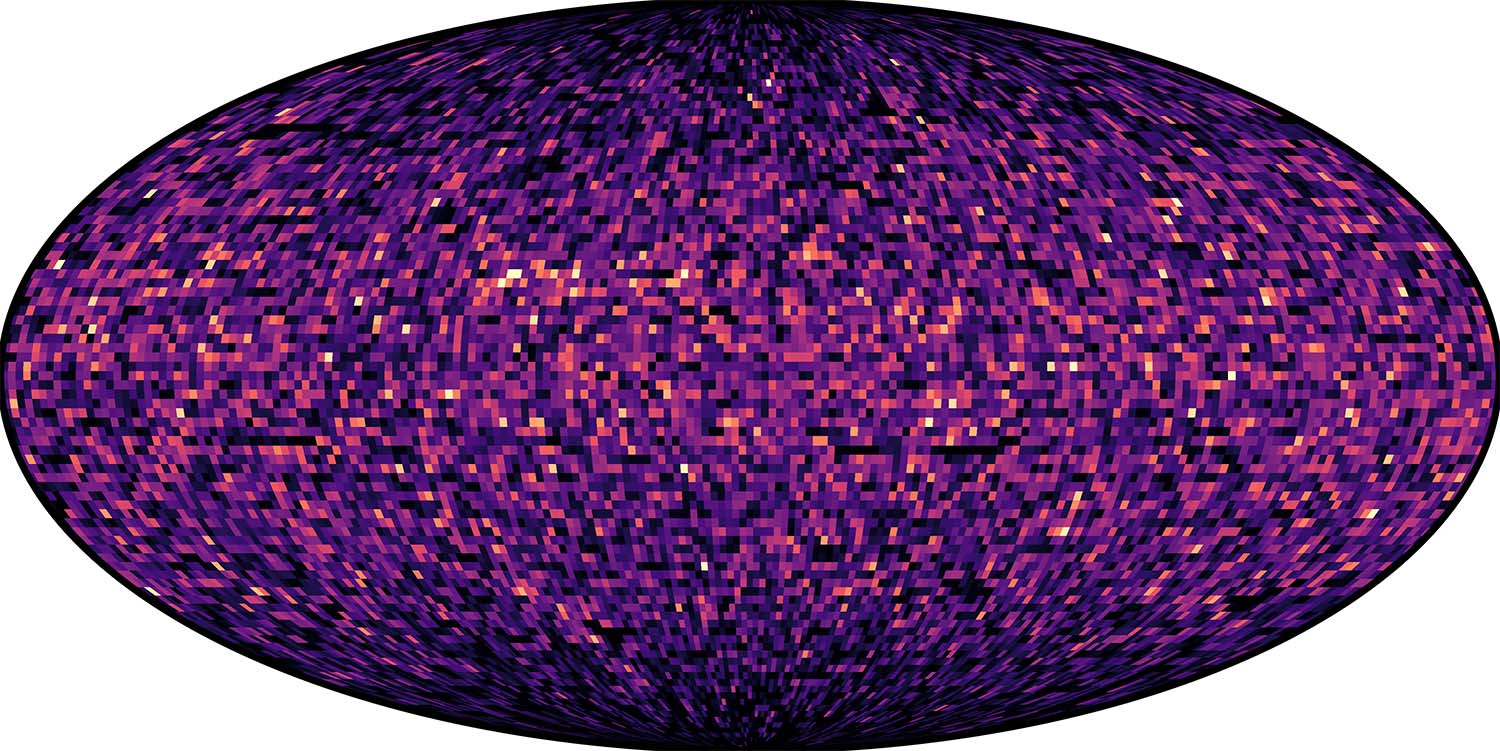} & 
        \includegraphics[width=0.14\textwidth]{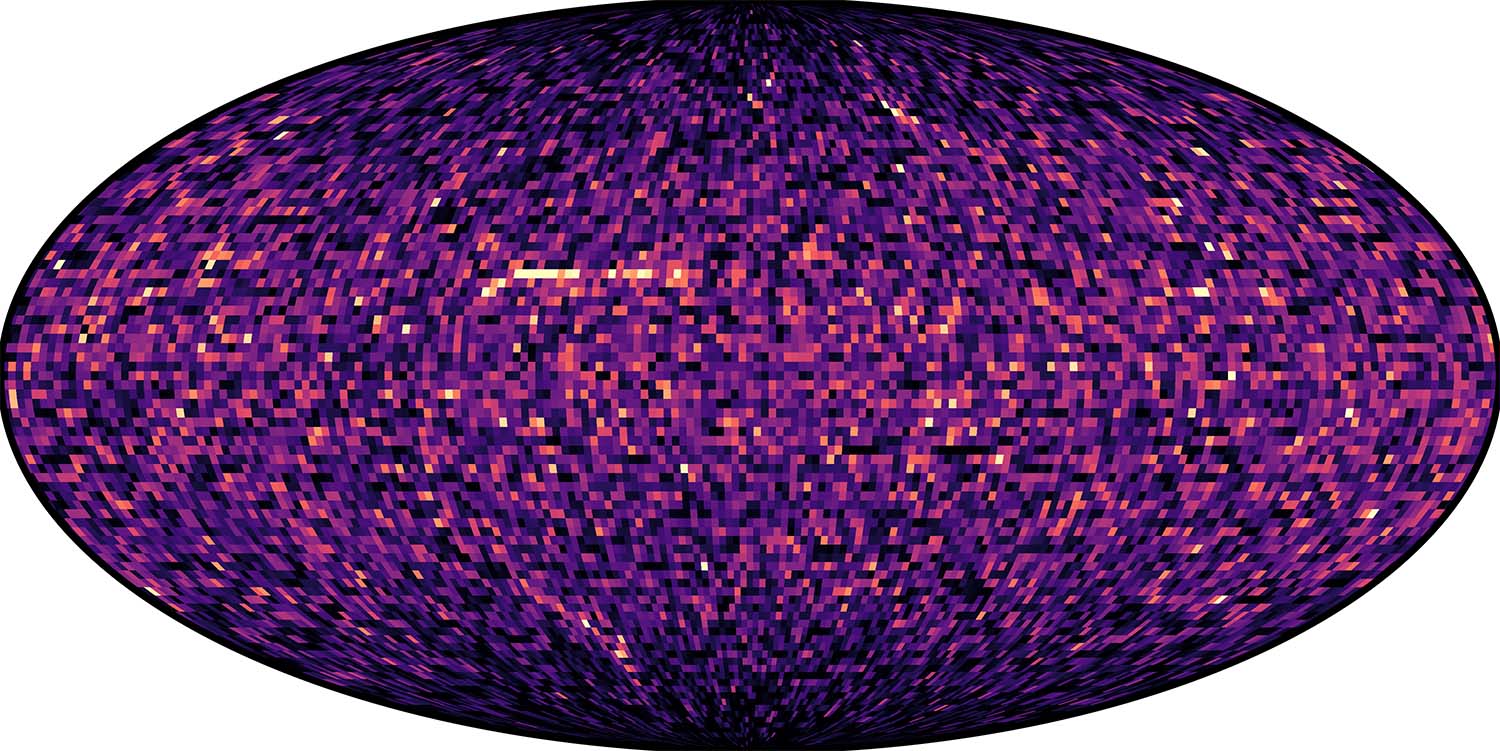} \\
    \end{tabular}
    \caption{CNFM on a manifold (sphere): the trained CNF $\phi_t$ is pushing noise $x\sim p_0$ to data $\phi_t(x)$ (top, from left $t=0$ to right $t=1$); and the reverse time CNF taking data $x\sim p_\data$ to noise $\phi_{1-t}(x)$ (bottom).}
    \label{fig:series}
\end{figure*}

\section{Matching CNF and target probability}
We start by considering a \emph{target} probability density path $p \in \mathfrak{P}(\gM)$. We will use the notation $p_t$ to denote the density at time $t$, namely, $p_t=p(t,\cdot)$.
%
In a typical target path $p$, $p_0$ is some simple prior distribution, \eg, a distribution representing pure noise, and $p_1$ approximates the unknown data distribution, denoted $p_\data$ and is practically approximated by some empirical set of samples. 

Our goal is to match $p$ and the density path $q\in \mathfrak{P}(\gM)$, generated by a CNF $\phi_t$ from the prior $p_0$. The CNF $\phi_t$ is defined by \eqref{e:phi} via a learnable time dependent vector field $v_\theta\in\mathfrak{X}(\gM)$, with parameters $\theta\in \Real^p$. In more detail, we define the CNF Matching (CNFM) problem as the following optimization problem:
\begin{subequations}\label{e:cnfm}
\begin{align}
\min_\theta &\quad \dist( p  \bb   q) \\ \label{e:cnfm_constaints}
 \mathrm{s.t.}&\quad  q_t=\phi_{t*}p_0, \quad  t\in[0,1]
\end{align}
\end{subequations}
where $\dist$ is a probability divergence between probability density \emph{paths}. That is, for density paths $p,q\in \mathfrak{P}(\gM)$, $\dist(p,q)\geq 0$, and $\dist(p,q)=0$ iff $p_t\equiv q_t$ for all $t\in [0,1]$.

Adapting existing CNF approaches to optimize \eqref{e:cnfm} would require evaluating $q_t$, which is provided only through solutions to an ODE (see also the discussion in Section \ref{ss:previous_work_cnf}), and will therefore introduce a substantial computational challenge. Instead, we construct a novel divergence $\dist$, called the Probability Path Divergence (PPD), that does not require sampling of $q$ or enforcing \eqref{e:cnfm_constaints} explicitly, and therefore sidesteps the need for solving ODE during training. Furthermore, we will show that PPD bounds standard probability divergences such as total variation, $\alpha$, and reverse KL. Figure \ref{fig:series} depicts an example of a CNF, $\phi_t$, trained with CNFM and PPD using a target path $p$ that is interpolating between uniform and checkerboard data over the sphere. In the top row we depict random uniform samples over the sphere $x\sim p_0$ (left) pushed by the CNF, \ie, $\phi_t(x)$, for several times $t\in[0,1]$, reaching the desired checkerboard distribution at $t=1$ (right). The bottom row shows the CNF pulling, \ie, $\phi_{1-t}(x)$, data samples $x\sim p_\data$ (left), reaching a uniform distribution at time $t=1$ (right). 

%

\subsection{Logarithmic Mass Conservation}
%
As a first step in constructing the PPD we derive a Partial Differential Equation (PDE) involving the log density path $\log p$ and a vector field $v$, such that it is satisfied iff the CNF $\phi_t$, defined by $v$, generates $p$. 
We name this equation the Logarithmic Mass Conservation (LMC) formula. 
\begin{theorem}\label{thm:lmc}
Consider a CNF $\phi_t:\gM\too\gM$ defined by a smooth, time dependent vector field $v\in \mathfrak{X}(\gM)$ as in \eqref{e:phi}, and a probability density path $p \in \mathfrak{P}(\gM)$. Then $p$ is generated by $\phi_t$, \ie,
\begin{equation}\label{e:LMC_phi_generates_alpha_t}
p_t = \phi_{t*} p_0, \quad \forall t\in [0,1]
\end{equation}
if and only if the LMC formula holds over $[0,1]\times\gM$:
\begin{equation}\label{e:lmc}
    \partial_t \log p_t + \ip{\nabla \log p_t , v} + \mathrm{div}(v) = 0
\end{equation}
\end{theorem}
The LMC formula can be proved with the aid of the mass conservation formula, also known as the continuity equation and equivalent to \eqref{e:lmc} \citep{villani2009optimal}:
\begin{equation}\label{e:mass_conserve}
\partial_t p_t + \mathrm{div}(p_t v) = 0,
\end{equation}
where $\mathrm{div}$ denotes the divergence operator over the manifold $\gM$. We assumed $p>0$ and therefore dividing both sides by $p_t$ leads to 
$$\frac{\partial_t p_t}{p_t}+\frac{\ip{\nabla p_t, v}+p_t \mathrm{div}(v)}{p_t}=0,$$
where we also used the fact that $\mathrm{div}(fv)=\ip{\nabla f, v} + f\mathrm{div}(v)$. Finally noting that $\partial_t \log p_t= \frac{\partial_t p_t}{p_t}$, and $\nabla_x \log p_t= \frac{\nabla_x p_t}{p_t}$ we get that \eqref{e:lmc} is equivalent to  \eqref{e:mass_conserve}. See Appendix \ref{a:proof_lmc} for more details. 

The benefit of using the LMC formula over the standard mass conservation formula is that it is formulated directly in terms of the log probability $\log p_t$, which reduces numerical issues for high dimensions.

\pagebreak
\subsection{Probability path divergence}
Plugging a fixed target path $p\in\mathfrak{P}(\gM)$ in the LMC formula (\eqref{e:lmc}) provides a necessary and sufficient condition for $v$ to generate $p$ via a CNF. Motivated by this observation, we define a family of probability path divergences (PPD), parameterized by an integer $\ell\geq 1$, comparing $p,q \in \mathfrak{P}(\gM)$ where $q_t=\phi_{t*}p_0$:
\begin{equation}\label{e:d}
      \dist_\ell(p \bb  q)\hspace{-2pt}=\hspace{-2pt}\E_{t,x\sim p_t} \Big | \partial_t \log p_t \hspace{-2pt}+\hspace{-2pt} \ip{\nabla \hspace{-2pt}\log p_t, v}\hspace{-2pt} +\hspace{-2pt} \mathrm{div}(v)\Big |^\ell\hspace{-2pt}
\end{equation}
where $t$ is distributed over $[0,1]$, \eg, uniform $t\sim \gU[0,1]$.  $\dist_\ell(p \bb q)\geq 0$ by construction, and Theorem \ref{thm:lmc} implies that $\dist_\ell(p \bb q)=0$ iff $p_t\equiv q_t$ for all $t\in [0,1]$. Using this path divergence in the CNFM problem (\eqref{e:cnfm}) we arrive to the following instantiation:
\begin{equation}\label{:cnmf_simple}
    \min_\theta \E_{t, x\sim p_t}  \Big | \partial_t \log p_t + \ip{\nabla \log p_t, v_\theta} + \mathrm{div}(v_\theta)\Big |^\ell
\end{equation}
where $v_\theta$ is the learnable vector field defining the CNF $\phi_t$ generating $q_t$. Importantly, evaluating the PPD $\dist_\ell(p\bb q)$ and its derivatives with respect to $\theta$ does not require access to $q$ and $\phi_t$, and therefore sidesteps solving the ODE in \eqref{e:phi} during training.


The following Theorem relates the path divergence $\dist_\ell$ to standard divergences of probability densities. We consider $f$-divergences \cite{ali1966general,csiszar1967information} of two probability densities $\mu,\nu$ defined by \begin{equation}\label{e:f_div}
    D_f(\mu \bb \nu) = \int_\gM f\parr{\frac{\mu(x)}{\nu(x)}}\nu(x)dV_x
\end{equation}
where $f:\Real_{\geq 0}\too\Real$ is a strictly convex function satisfying $f(1)=0$. $f$-divergences satisfy the standard statistical divergence properties: $D_f(\mu \bb \nu)\geq 0$, and $D_f(\mu \bb \nu)=0$ iff $\mu\equiv \nu$. $f$-divergences generalize standard divergences such as KL (with the choice $f(t)=t\log t$), reverse KL ($f(t)=-\log t$), total variation ($f(t)=\abs{t-1}$), and $\alpha$-divergences ($f(t)=1-t^\alpha$ with $\alpha\ne 1,0$).  
We prove: 
\begin{theorem}\label{thm:relation_to_divergence}
Consider paths $p,q\in \mathfrak{P}(\gM)$ where $q$ is generated by a CNF $\phi_t:\gM\too\gM$, and $q_0=p_0$. Then for all $T\in [0,1]$
\begin{equation}\label{e:bound}
    \dist_\ell(p \bb q)^{\frac{1}{\ell}} \geq D_f(p_T \bb q_T)
\end{equation}
where 
\begin{equation*}
f(t)=
 \begin{cases}
      \abs{t-1} &\ell = 1 \qquad \qquad \text{(total variation)}\\
      \ell\parr{1-t^{\frac{1}{\ell}}} & 1 < \ell < \infty \qquad \qquad (\alpha) \\
      -\log t & \ell=\infty \qquad \qquad \text{(reverse KL)}
     \end{cases}    
\end{equation*}
\end{theorem}

\begin{wrapfigure}{r}{0.35\columnwidth}\hspace{-10pt}
    \includegraphics[width=0.33\columnwidth]{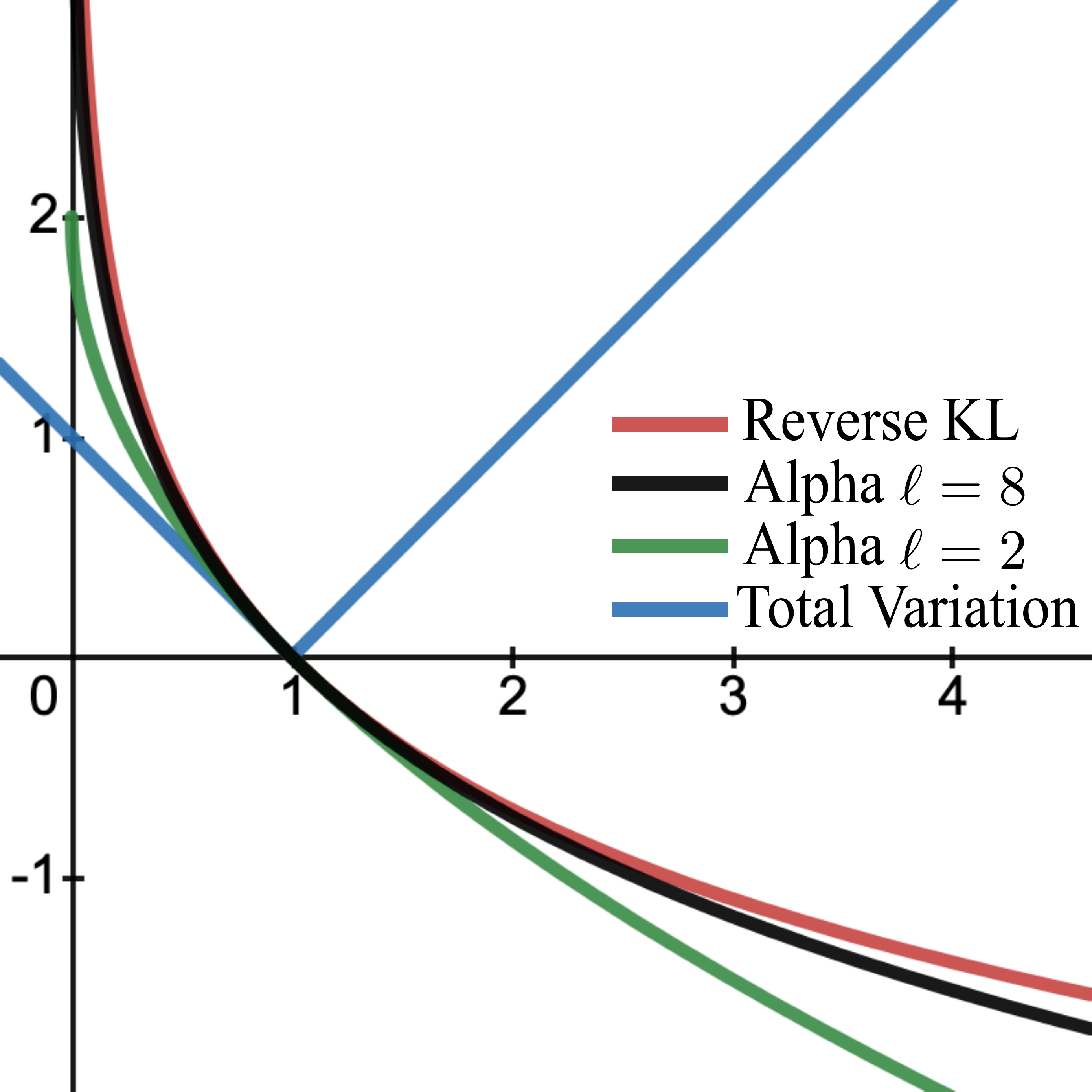}\vspace{-12pt}
  \caption{$f$ instances, see  Theorem~\ref{thm:relation_to_divergence}.}\label{fig:f}
\end{wrapfigure}

Theorem \ref{thm:relation_to_divergence} shows that the path divergence $\dist_\ell$ bounds the respective $f$-divergences of $p_T$ and $q_T$ for all times $T\in[0,1]$. Figure \ref{fig:f} visualizes four instances of $f$ corresponding to different choices of $\ell$.
Note, that with the exception of $\ell=1$, all $f$ are differentiable and have the same derivative at $1$, which means they have similar value and derivatives when evaluating the divergence of nearby probability densities. As $\ell \too \infty$ we can see the $f$ functions gets close to the $-\log t$ limit. 


Specifically, in the $\ell=\infty$ case of Theorem \ref{thm:relation_to_divergence}, we mean that the inequality \eqref{e:bound} holds in the limit as $\ell\too \infty$, or more precisely, \begin{equation*}
    \liminf_{\ell\too\infty}\dist_\ell(p\bb q)^{1/\ell}\geq D_f(p_T\bb q_T),
\end{equation*} where we also assume that $D_f(p_T\bb q_T)<\infty$. 
To prove this theorem we will use the following lemma, proved in Appendix \ref{a:proof_d_form}.
\begin{lemma}\label{lem:d_form}
Consider paths $p,q\in \mathfrak{P}(\gM)$ where $q$ is generated by a CNF $\phi_t:\gM\too\gM$, and $q_0=p_0$. Then the following holds:
\begin{equation*}
    \dist_\ell(p \bb q) = \E_{x\sim p_0} \int_0^1 \frac{p_t(\phi_t(x))}{q_t(\phi_t(x))}\abs{\partial_t \brac{\log \frac{p_t(\phi_t(x))}{q_t(\phi_t(x))}}}^\ell dt
\end{equation*}
\end{lemma}
We now use Lemma \ref{lem:d_form} to prove each case of Theorem \ref{thm:relation_to_divergence}:
\paragraph{$\ell=1$ case.} For $\ell=1$,  Lemma \ref{lem:d_form} provides the following form for $\dist_1$:
\begin{equation}
    \dist_1(p \bb q) = \E_{x\sim p_0} \int_0^1  \abs{\partial_t  \brac{\frac{p_t(\phi_t(x))}{q_t(\phi_t(x))}}} dt
\end{equation}
which shows that for $\ell=1$ the path divergence is equivalent to the Total Variation norm of the density ratio $p_t/q_t$ along trajectories of the flow. Second, Jensen's inequality with the convex function $\abs{\cdot}$ provides for every $T\in[0,1]$
\begin{align*}
    \dist_1(p \bb q)&\geq \E_{x\sim p_0} \int_0^T  \abs{\partial_t \brac{ \frac{p_t(\phi_t(x))}{q_t(\phi_t(x))}}} dt \\ 
    &\geq \E_{x\sim p_0} \abs{\frac{p_T(\phi_T(x))}{q_T(\phi_T(x))} - 1  } = \E_{x\sim q_T}\abs{\frac{p_T(x)}{q_T(x)} - 1  } \\ &= D_{f}(p_T,q_T)
\end{align*}
where the first inequality is due to the fact that we integrate over the smaller interval $[0,T]$, in the first equality we used the fact that $\phi_T(x)\sim q_T$ if $x\sim p_0$, and in the last equality we took $f(t)=\abs{t-1}$. 

\paragraph{$1<\ell<\infty$ case.}
Lemma \ref{lem:d_form} again with Jensen's inequality of the convex function $\abs{\cdot}^\ell$ provides
\begin{align} \nonumber
    \dist_\ell(p \bb q)^{\frac{1}{\ell}} &\geq \abs{\E_{x\sim p_0}  \int_0^T\hspace{-2pt} \brac{\frac{p_t(\phi_t(x))}{q_t(\phi_t(x))}}^{\frac{1}{\ell}}\hspace{-4pt} \partial_t\hspace{-3pt} \brac{ \hspace{-2pt}\log \frac{p_t(\phi_t(x))}{q_t(\phi_t(x))} \hspace{-2pt}}\hspace{-2pt} dt } \\ \nonumber
    &=  \abs{\E_{x\sim p_0}  \int_0^T \ell \partial_t \brac{\frac{p_t(\phi_t(x))}{q_t(\phi_t(x))}}^{\frac{1}{\ell}} dt }\\ \nonumber
    &=\abs{\E_{x\sim p_0}  \ell\parr{\brac{\frac{p_T(\phi_T(x))}{q_T(\phi_T(x))}}^{\frac{1}{\ell}} - 1} } \\ \label{e:before_lim}
    &=\abs{\E_{x\sim q_T}  \ell\parr{\brac{\frac{p_T(x)}{q_T(x)}}^{\frac{1}{\ell}} - 1} } \\ \nonumber
    & = D_f(p_T \bb q_T) 
\end{align}
with $f(t)=\ell(1-t^\frac{1}{\ell})$. 
\paragraph{$\ell=\infty$ case.}
First, we note that for any $t>0$, $ \ell(1-t^{\frac{1}{\ell}})\nearrow -\log(t)$, that is, $\ell(1-t^{\frac{1}{\ell}})$ is monotonically increasing and converging to $-\log (t)$ as $\ell\too\infty$ (see Appendix \ref{a:relation_to_divergence} for a proof). Next, consider \eqref{e:before_lim} and move to the limit $\ell\too\infty$:
\begin{align} \nonumber
    \liminf_{\ell\too\infty}\dist_\ell(p \bb q)^{\frac{1}{\ell}} &\geq \lim_{\ell\too\infty}\abs{ \E_{x\sim q_T}  \ell\parr{1-\brac{\frac{p_T(x)}{q_T(x)}}^{\frac{1}{\ell}} }} \\ \label{e:lim_switch} &= -\E_{x\sim q_T} \log \frac{p_T(x)}{q_T(x)}\\ \nonumber
    &= D_f(p_T \bb q_T)
\end{align}
with $f(t)=-\log(t)$. 
The previous to last equality (integral and limit switch) is justified in Appendix \ref{a:relation_to_divergence}; the minus sign is due to the fact that $D_f(p_T \bb q_T)\geq 0$.

\subsection{Target paths}\label{ss:target_path}
The last ingredient needed for defining the probability path divergence (\eqref{e:d}) is the target path $p\in\mathfrak{P}(\gM)$. In our framework, $p$ should be defined satisfying the following requirement: 
\begin{enumerate}[(i)]
    \item $p_0$ is pure noise, \eg, a standard Gaussian or uniform.
    \item $p_1$ approximates the unknown data distribution $p_\data$.
    \item We have an efficient generation procedure for $x\sim p_t$.
    \item We have an approximation procedure for the time ($\partial_t$) and space ($\partial_x$) derivatives of $\log p_t(x)$.
\end{enumerate}
Note that these requirements do not mean we know of an SDE,  generating random variables distributed as $p_t$, nor a PDE (Fokker-Planck) with $p_t$ as its solution. In fact, below we construct paths for which an SDE/PDE characterization is not known. In that context the target paths we consider are general; see discussion in Section \ref{ss:previous_works_sde}. 

In the following we construct target paths $p\in\mathfrak{P}(\gM)$ for several manifolds of interest. At the base of our construction is a kernel $p_\tau(x|y)$, namely a probability density in $x\in\gM$, centered at $y\in\gM$, with scale $\tau>0$. We define our (ideal) target path $p\in\mathfrak{P}(\gM)$ by
\begin{equation}\label{e:path_ideal}
    p_t(x) = \int_\gM p_\tau(x\,|\,\gamma_t(y))p_{\data}(y)dV_y
\end{equation}
where $\tau=\tau(t)$, $t\in [0,1]$, is a time-dependent scale function, and $\gamma:[0,1]\times\gM\too\gM$ is some differentiable in $t$ map. In practice we don't know $p_\data$, rather, we have an empirical sample  $\set{y_i}_{i=1}^m$, drawn i.i.d.~from $p_\data$. Therefore we use the following approximation of \eqref{e:path_ideal} 
\begin{equation}\label{e:path}
    p_t(x) = \frac{1}{m}\sum_{i=1}^m p_\tau(x\,|\,\gamma_t(y_i))
\end{equation}
Note, that if we know how to compute or approximate $\log p_\tau(x|\gamma_t(y_i))$ then $\log p_t(x)$, required for the computation of the PPD, has the form
\begin{equation*}
    \log p_t(x) \hspace{-2pt}=\hspace{-2pt} { \texttt{logsumexp}}\hspace{-2pt}\set{\log p_\tau(x|\gamma_t(y_i)}_{i=1}^m\hspace{-2pt}-\hspace{-1pt}\log m
\end{equation*} 
Depending on the type of manifold, we consider two basic target path constructions that differ in their prior $p_0$: \emph{Unimodal}, where the prior probability $p_0$ is centered around a single designated point in $\gM$. Unimodal prior is mainly suitable to non-compact manifolds with infinite volume such as Euclidean or hyperbolic spaces. \emph{Uniform}, where the prior $p_0$ is the uniform density over $\gM$. A uniform prior is suitable to compact manifolds such as spheres.  
%

\paragraph{Unimodal prior.}
Let $o\in \gM$ be some designated point, $\sigma_0,\sigma_1\geq 0$ initial and target scales. We define $p$ according to \eqref{e:path} by making the choices: 
\begin{equation}\label{e:single_mode_gamma}
    \gamma_t(y) = \exp_{o}(t\log_o y)\ \ , \ \ \sigma(t) = \sigma_0^{1-t}\sigma_1^t
\end{equation}
where $\tau=\sigma$ is the scaling function, and $\gamma_t(y)$ moves $y$ to the center $o$ along a geodesic (we assume the Riemannian $\exp_o,\log_o$ are defined in a  sufficiently large neighborhood of $o$ and $T_o\gM$).
With these choices, $p_t$ starts with a single mode density $p_0$, centered at $o\in\gM$, and then splits the unit mass, moving each $\frac{1}{m}$ part towards the empirical sample $y_i$ along a geodesic while concentrating the density. 

\emph{Euclidean.} Let us instantiate the unimodal path for the Euclidean space,  $\gM=\Real^d$, with the standard metric $\ip{v,u}=v^Tu$, where $v,u\in\Real^d$ are (always) column vectors. Our kernel in this case is the Gaussian, $p_\sigma(x|y)=\gN(x|y,\sigma^2 \mI)$ with mean $y\in\Real^d$ and covariance $\sigma^2 \mI$. Furthermore, for $o\in\Real^d$, $\exp_o(t\log_o y_i) = o+t(y_i-o) = (1-t)o+t y_i$. Therefore, \eqref{e:path} takes the form
\begin{equation}\label{e:path_euc}
    p_t(x) = \frac{1}{m}\sum_{i=1}^m \gN(x\,|\,(1-t)o+ty_i, \sigma^2 \mI)
\end{equation}
and we take $\sigma_0=1$ to represent a standard Gaussian prior, \ie, $\sigma(t)=\sigma_1^t$, and $\sigma_1>0$ is the (only)  hyper-parameter. 


\paragraph{Uniform prior.}
In this family of paths we consider compact manifolds $\gM$ and start from the uniform density $p_0$.
%
We assume in this case we have a kernel $p_\kappa(x|y)$ such that there exists a finite $\kappa_0\geq 0$, for which $p_{\kappa_0}(x|y)\equiv |\gM|^{-1}$ for all $y\in\gM$, \ie, $p_{\kappa_0}$ represents the uniform density. One way to construct such a kernel on compact submanifolds of $\Real^{d+1}$, $\gM\subset\Real^{d+1}$, is by restricting an Euclidean Gaussian in $\Real^{d+1}$ to $\gM$; we discuss such a construction on the sphere below. In this case we define the target path using \eqref{e:path} again by making the choices 
\begin{equation}
    \gamma_t(y)=y \ \ , \ \ \kappa(t)=(1-\kappa_0 + \kappa_1)^{t} + \kappa_0 - 1
\end{equation}
where $\tau=\kappa$ is the scaling function, and $\gamma_t(y)$ leaves samples at their original location. 

\emph{Sphere.} We instantiate the uniform prior paths to the unit spheres $\gM=\gS^d\subset \Real^{d+1}$ with the induced metric from the Euclidean $\Real^{d+1}$. The von Mises-Fisher (vMF) kernel \cite{mardia2014statistics} is:
\begin{equation}\label{e:vmf}
    p_\kappa(x|y) = c_d(\kappa) \exp(\kappa x^Ty),
\end{equation} where $c_d(\kappa)$ is the normalization constant detailed in Appendix \ref{a:constant_of_vmf}. vMF can be seen as a restricted Gaussian $\exp(-\kappa \norm{x-y}^2_2)$ to the unit sphere $x,y\in\gS^d$ with the relevant normalization constant. For $\kappa=0$, $p_0(x|y)$ is uniform over the sphere for all $y\in \gS^d$. Hence we take $\kappa_0=0$, which leaves $\kappa(t)=(1+\kappa_1)^t-1$, and $\kappa_1>0$ is the (only) hyper-parameter in this case. The target path takes the form
\begin{equation}\label{e:path_sphere}
    p_t(x) = \frac{1}{m}\sum_{i=1}^m p_\kappa (x| y_i)
\end{equation}

\paragraph{Paths on products of manifolds.} We conclude the section with generalizing the target path construction to product of manifolds.  Let $\gM = \gM^1 \times \ldots \times \gM^N$.
Each point $x\in \gM$ is represented as a tuple $x=(x^1,\ldots,x^N)$, where $x^j\in \gM^j$.  For example, in robotics, a robot's state can be represented by the sequence of locations and/or rotations of its joints, \ie, each $\gM^j$ is either a sphere ($\gS^3$ for 3D rotations represented as quaternions; $\gS^1$ for 2D rotations) or an Euclidean space (representing positions).  Let $p_{\tau^j}$ be a kernel defined in $\gM^j$, and $\gamma^j$ is a deformation of $\gM^j$. For example, for $\gM^j$ being the Euclidean plane or a sphere we can use the above definitions for kernels $p_{\tau^j}$. 
Let $\set{y_i}_{i=1}^m\subset\gM$ be i.i.d.~samples from $p_\data$ over $\gM$. We define the kernel for $\gM$ by
\begin{equation}\label{e:product_kernel}
    p_{\tau}(x|y_i) = \prod_{j=1}^N p_{\tau^j}(x^j | \gamma^j_t(y^j_i) )
\end{equation}
We note that $p_\tau(x|y)$ is a probability density in $x\in \gM$, and if $p_{\tau^j}(x^j | y^j)$ is concentrated (as a function of $x^j\in\gM^j$) around $y^j$ for all $j$, then $p_\tau(x|y)$ is concentrated (as a function of $x\in \gM$) around $y$. Lastly, and use \eqref{e:path} again to define out target path $p\in \mathfrak{P}(\gM)$. Further implementation details for the vector field $v_\theta$ are in Appendix~\ref{a:vector_field_rep}.



\section{Previous works}

\subsection{Relations to existing CNF models} 
\label{ss:previous_work_cnf}
The LMC formula (\eqref{e:lmc}) is a linear first order PDE in $\log p_t$. Solving it using the method of characteristics \citep{evans1997partial} provides a simple proof of the Instantaneous Change of Variables Theorem from \citep{chen2018neural} and generalizes it to the manifold setting. Indeed, using the chain rule and the LMC we have \begin{align} \nonumber
    \partial_t \brac{\log q_t (\phi_t)} &= \partial_t \log q_t (\phi_t) + \ip{\nabla_x \log q_t (\phi_t) , v(t,\phi_t)} \\ \label{e:instant}
    & = -\mathrm{div}\, v(t,\phi_t)
\end{align}
where $\partial_t \brac{\log q_t (\phi_t)}$ denotes the total derivative w.r.t.~$t$.
Training a neural ODE by maximizing the  likelihood of the data points $x_i\in\gM$ entails computing $\log q_t(x_i)$ and its derivatives w.r.t.~the parameters of the vector field $v_t$. Using the characteristic method \citep{chen2018neural,lou2020neural,mathieu2020riemannian,falorsi2020Neural} this amounts to solving an ODE for $(\log q_t(\phi_t), \phi_t)$ (equations \ref{e:phi} and \ref{e:instant}) and differentiating the solution (which involves another ODE solve). In contrast, minimizing the PPD does not require solving an ODE during training.  

Moser Flow (MF) \citep{rozen2021moser} suggests to train a CNF by formulating the model density as $q_1 = p_0 - \mathrm{div}(u)$, where $u$ is time independent vector field over $\gM$. Its relation to our method can be seen by making the choice $q_t = (1-t)p_0 + tq_1$, where $p_0$ and $q_1$ are prior and model probability densities, respectively. Indeed, plugging this path in the mass conservation equation (equation \ref{e:mass_conserve}) gives
\begin{equation*}
    q_1- p_0 + \mathrm{div}\parr{q_t v} =0
\end{equation*}
\pagebreak
which directly leads to MF by plugging $u=p_t v$ as a time independent solution to this equation. Although MF also avoids solving an ODE during training and generalizes to manifolds, it incorporates an additional loss term for keeping the model density $q_1$ positive; this loss term has high variance and does not scale to high dimensions. Furthermore, MF models probabilities rather than log-probabilities, which also hinders modeling high dimensional densities. Lastly, MF models a particular probability path (convex combinations of prior and model), while our framework can match more general paths. 


\subsection{Relations to diffusion and score based generative models}
\label{ss:previous_works_sde}



Another body of related work concerns diffusion-based generative models \cite{sohl2015deep, ho2020denoising} and SDE/score-based generative models~\cite{song2019generative,song2020score}.
Both approaches also use a certain probability density path, called the \emph{forward process}, to train their generative model. The forward process is a (continuous or discrete) time dependent noising scheme converting the data distribution to a simple, easy to sample from, prior distribution.  In diffusion models the forward process is defined by a markov chain, whereas for score models it is defined by an SDE. The forward process is used for training the \emph{reverse process} parameterized with a neural network. The reverse process is used to generate samples from the prior distributions. 

Training diffusion/score models entails: (i) sampling the forward process at arbitrary times $t$; this requires either a closed-form solution of the respective diffusion/SDE forward process (especially challenging over manifolds, more on this below), or simulating the process from time $t=0$ (costly). (ii) Spatial derivatives of the log transition kernel. Where the transition kernel is the probability of sampling a point $x$ at time $t$ from the forward process given an initial point $y\sim p_{\mathrm{data}}$. (iii) Known form of the reverse diffusion/SDE process.

Our CNFM approach, based on the LMC formulation, does not require the probability density path $p_t$ to be a known solution of a particular diffusion or SDE process and the reverse process is trivially obtained by solving the ODE in reversed time with the learned $v_\theta$. This makes the path choice in our approach more flexible compared to diffusion, score and SDE models, which are restricted to probability density paths defined by known diffusion processes (\eg, Gaussian) or SDEs with closed form transition kernels.  

This flexibility becomes especially important when the domain we want to learn on is not Euclidean. For example, consider the arguably simplest SDEs, describing Brownian motion over $\gM$. The corresponding probability kernel $p_t(x|y)$ is the fundamental solution to the heat equation $\partial_t p = \Delta p$, where $\Delta$ is the Laplace-Beltrami operator on the manifold $\gM$. Solutions to the heat equation are known in very few cases \cite{pennec2006intrinsic}, and even for the sphere the solution is only known as an infinite series of Legendre polynomials \cite{tulovsky2001formula}. Therefore using the SDE framework on manifolds will often require some numerical solutions to the relevant SDE/ODE.
In contrast, our LMC-based formulation provides the flexibility to specify arbitrary target probability paths between the prior and data densities. On the sphere for example, we use closed form paths defined by vMF distributions. 
%
For sampling, solving an ODE is generally easier than solving an SDE as ODE solvers have higher asymptotic convergence rates. For example, Euler's method has order 1 for ODE and only 0.5 for SDE \cite{kloeden2012numerical}. Furthermore, ODEs have simple higher order solvers like Runga-Kutta methods~\cite{dormand1980family} with widely used open-source implementations.

\section{Experiments}

We have tested the CNFM framework with the PPD for training CNFs on low and moderately high dimensional manifold data. In all experiments we generate the target path $p$ according to Section \ref{ss:target_path} with input data samples $\set{y_i}_{i=1}^m\subset \gM$. In all experiments we iterate over the dataset where the set $\set{y_i}_{i=1}^m\subset \gM$, which is used for the approximation of \eqref{e:path}, is simply the batch, that is $m=\mathrm{batch \;size}$. Note that for better approximation of \eqref{e:path}, we could take $m>\mathrm{batch \; size}$ and evaluate the loss only at a subset of size $\mathrm{batch \; size}$. Since the loss is still evaluated at only $\mathrm{batch\;size}$ of samples, \ie, the forward and backward costs are the same, and memory usage will not increase significantly. In general, we have found CNFM to facilitate faster training of CNFs with larger models, often producing state of the art sampling and density estimation.

\begin{figure}
    \centering
    \begin{tabular}{@{\hspace{1pt}}c@{\hspace{2pt}}c@{\hspace{2pt}}c@{\hspace{2pt}}c@{\hspace{2pt}}c@{\hspace{2pt}}c@{\hspace{1pt}}}
         \includegraphics[width=0.16\columnwidth]{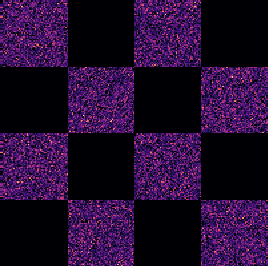}&  \includegraphics[width=0.16\columnwidth]{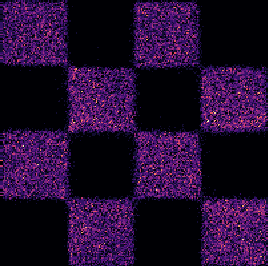} &
         \includegraphics[width=0.16\columnwidth]{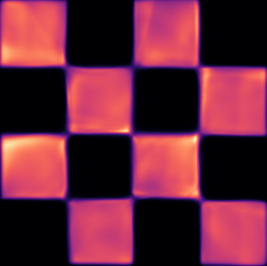}&  \includegraphics[width=0.16\columnwidth]{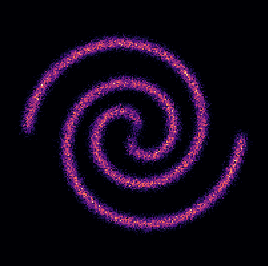} &
         \includegraphics[width=0.16\columnwidth]{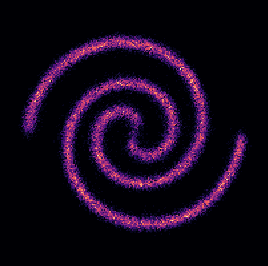}&  \includegraphics[width=0.16\columnwidth]{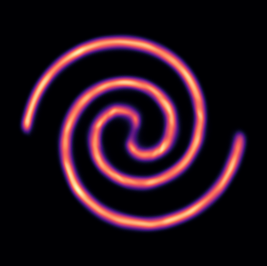} \\
         \includegraphics[width=0.16\columnwidth]{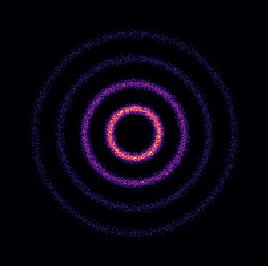}&  \includegraphics[width=0.16\columnwidth]{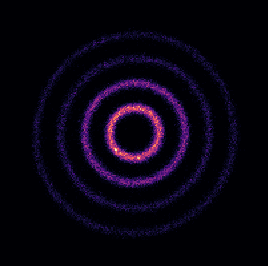} &
         \includegraphics[width=0.16\columnwidth]{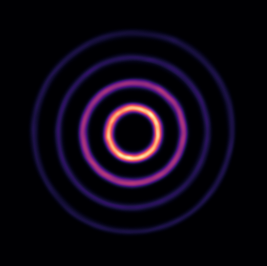}&  \includegraphics[width=0.16\columnwidth]{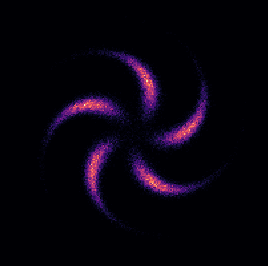} &
         \includegraphics[width=0.16\columnwidth]{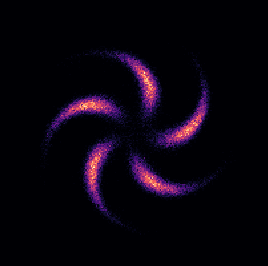}&  \includegraphics[width=0.16\columnwidth]{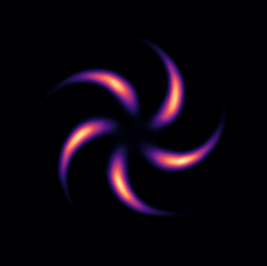} \\
         \includegraphics[width=0.16\columnwidth]{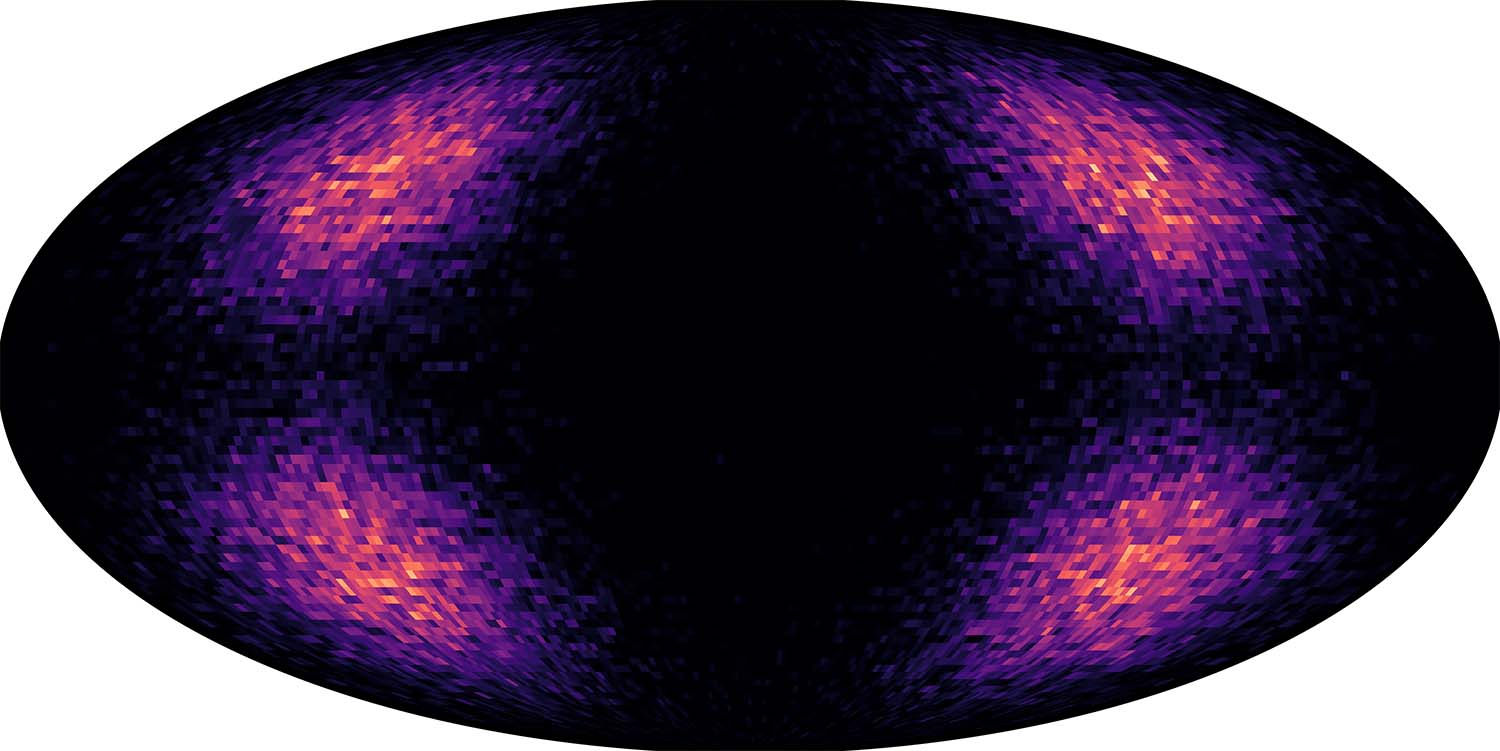}&  \includegraphics[width=0.16\columnwidth]{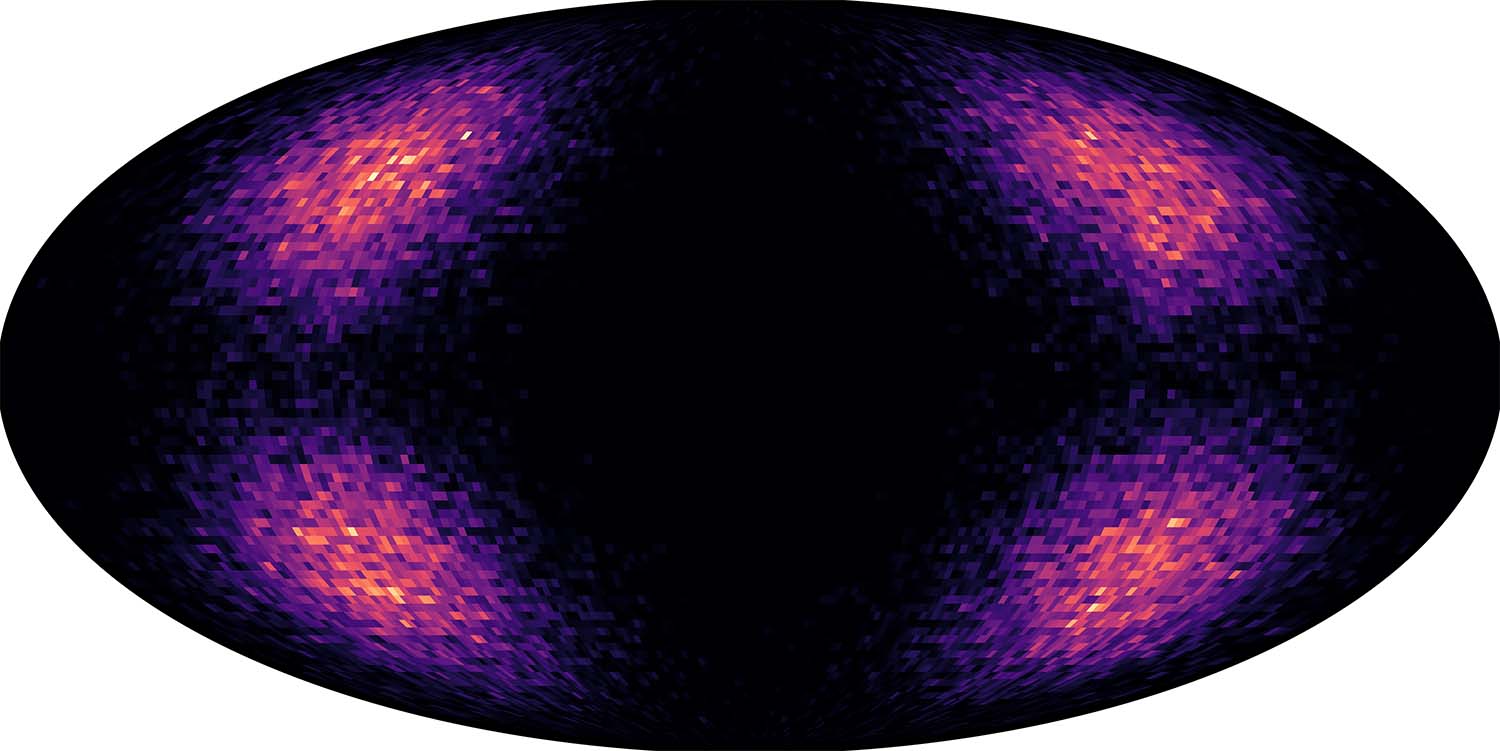}& 
         \includegraphics[width=0.16\columnwidth]{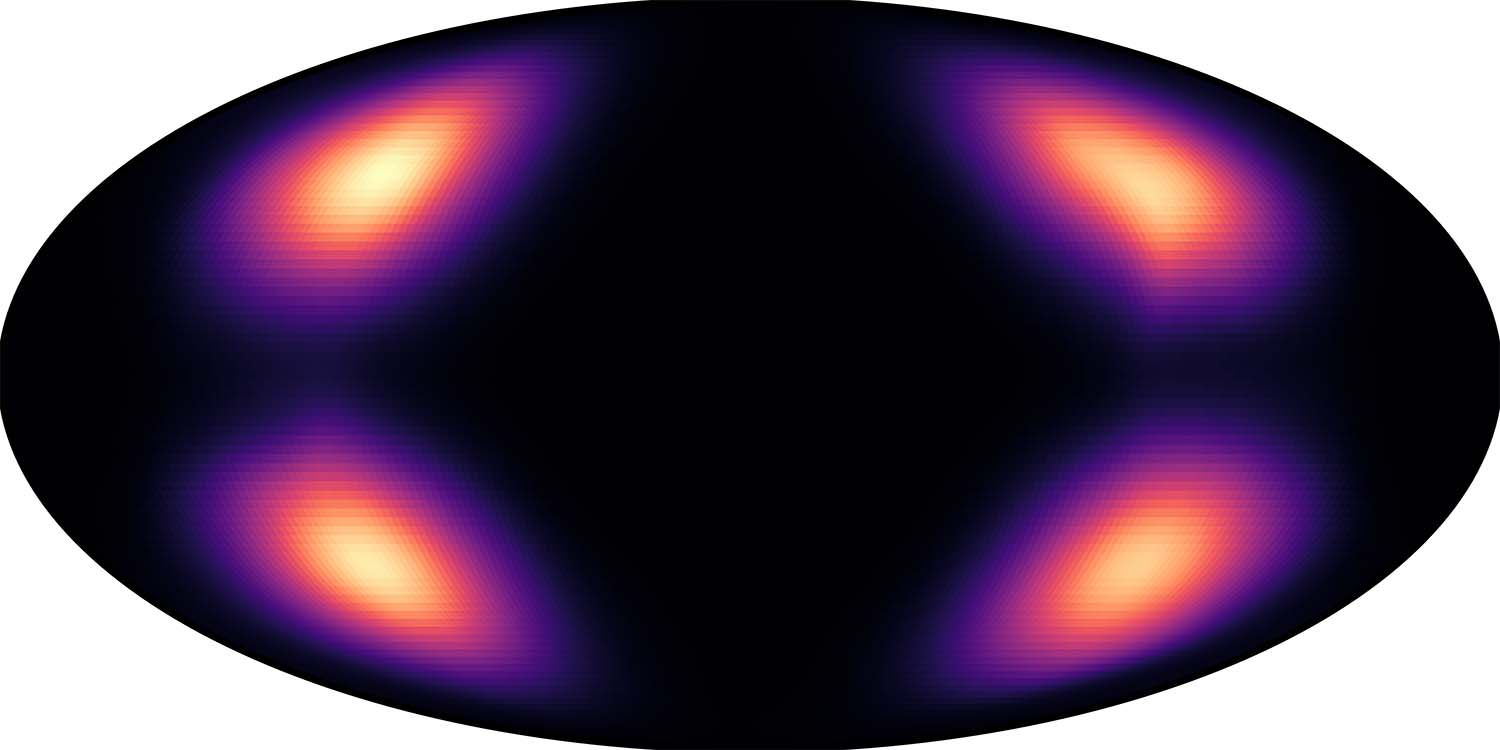}&  \includegraphics[width=0.16\columnwidth]{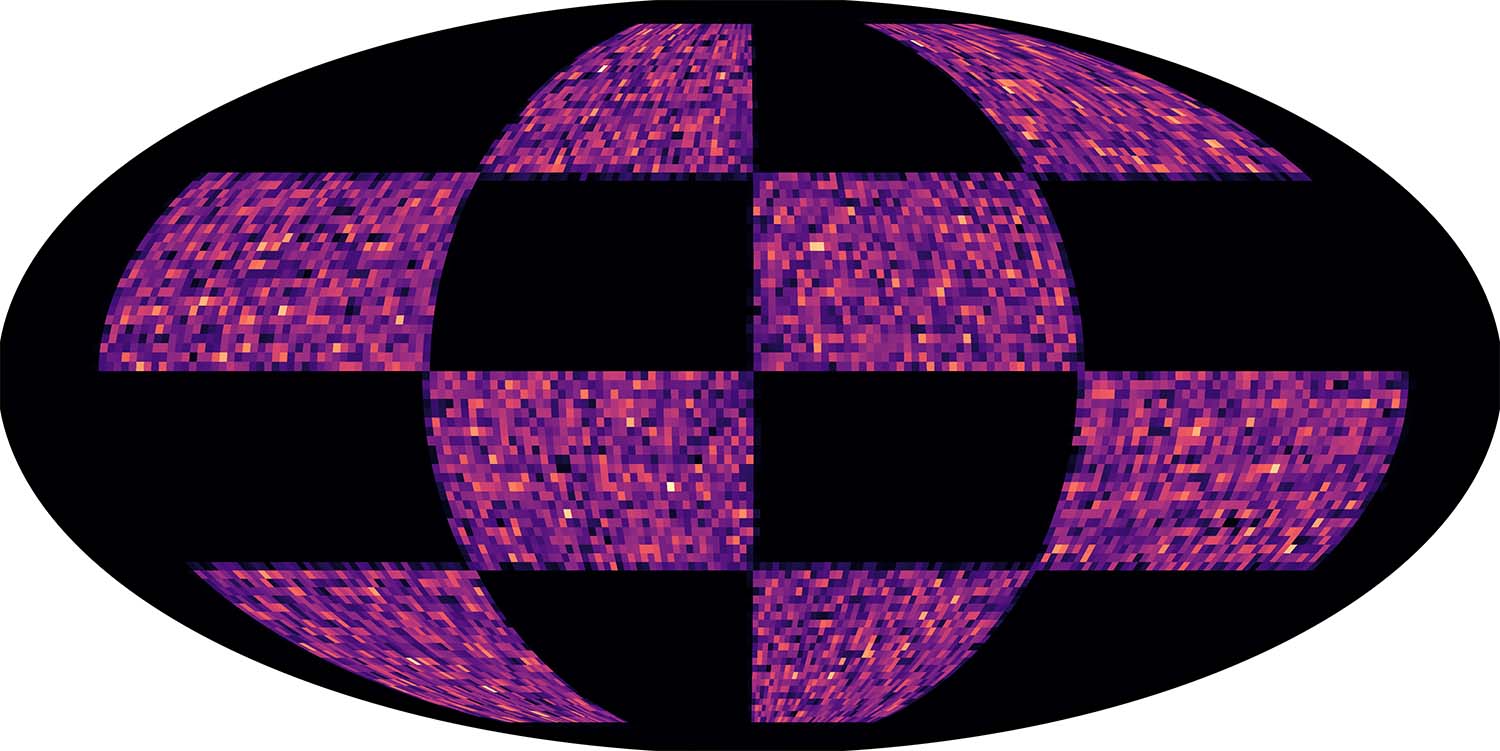}& 
         \includegraphics[width=0.16\columnwidth]{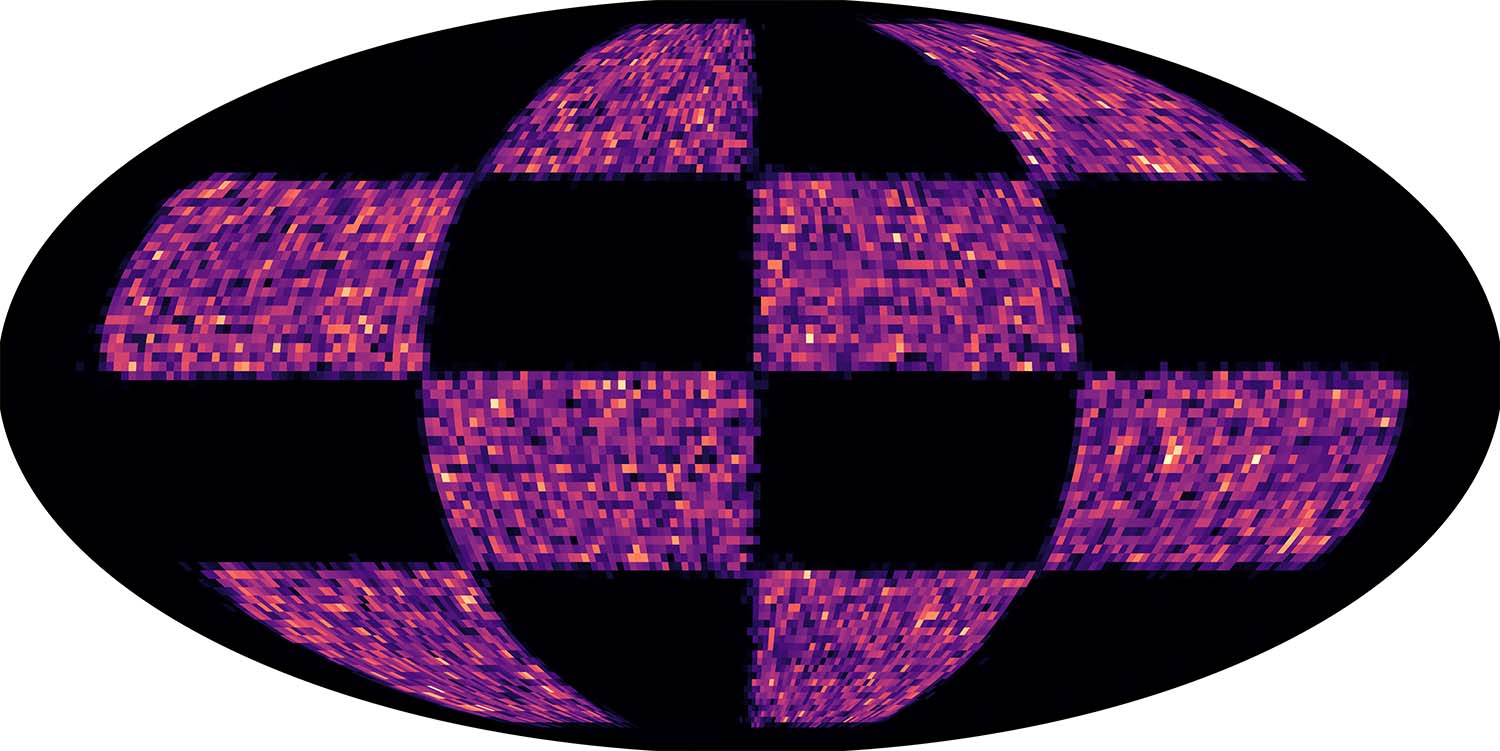}&  \includegraphics[width=0.16\columnwidth]{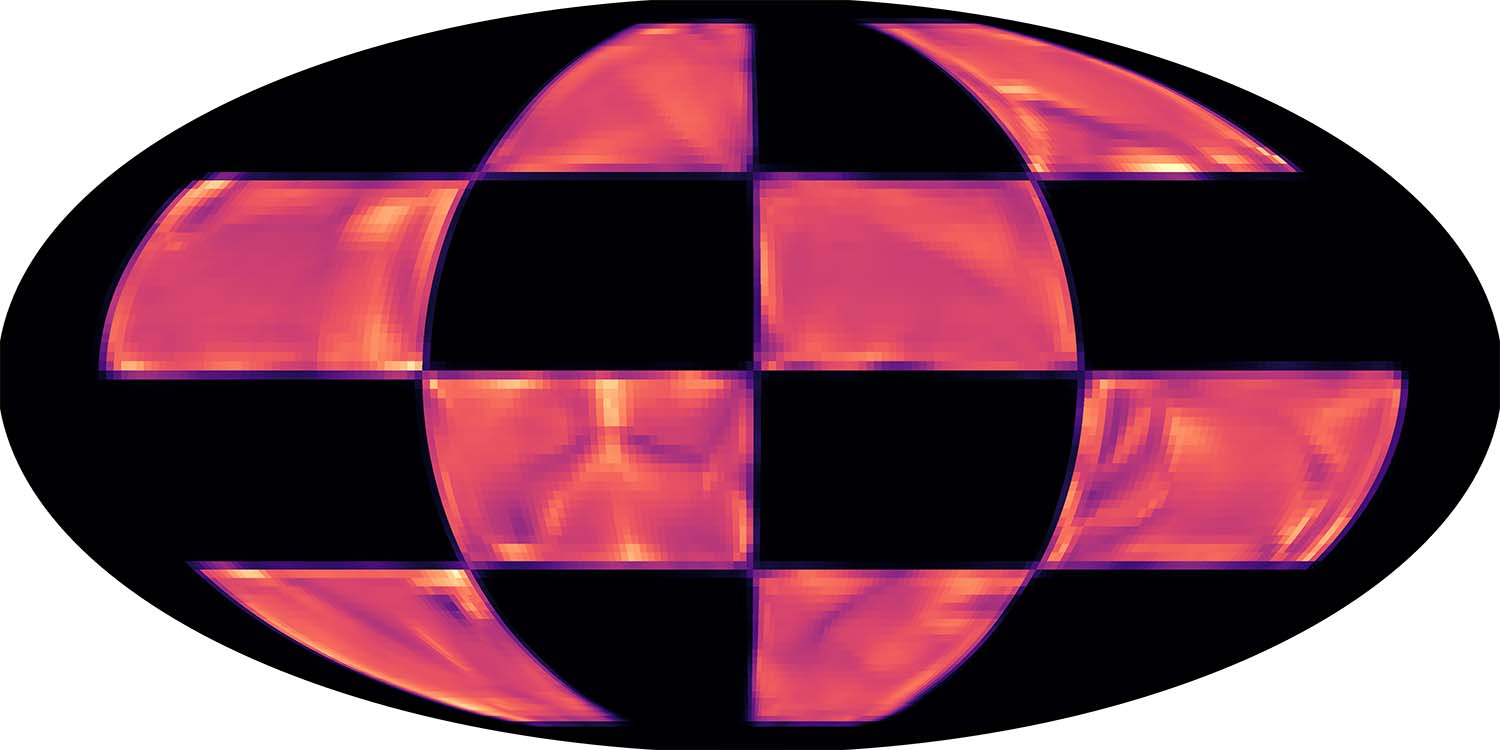}\\
    \end{tabular}
    \caption{2D toy densities. Each triplet shows (left to right): data samples, generate samples $x\sim q_1$, and learned model density $q_1$.}\vspace{-0pt}
    \label{fig:2d_toy}
\end{figure}

\subsection{Toy densities on $\Real^2$ and $\gS^2$}
In the first experiment we worked with samples drawn from standard toy distributions on the 2D Euclidean plane and sphere. For the Euclidean data we used the target path $p$ defined in \eqref{e:path_euc} with $p_0\sim \gN(x|0,\mI)$, the standard normal distribution, and $\sigma_1=0.01$. For the spherical data we used the target path $p$ as defined in \eqref{e:path_sphere} with $\kappa_1=5000$. We used MLP of $3$ layers of $256$ neurons for the $\Real^2$ data, and $6$ layers of $512$ neurons for $\gS^2$. We used PPD with $\ell=1$. Figure \ref{fig:2d_toy} depicts the data samples $y_i$ along side samples generated from the learned model, and the model densities. Note the high similarity between the learned and GT densities; for sphere visualizations we use Mollweide projection.

\subsection{Earth and climate dataset}
\begin{figure}
    \centering
    \begin{tabular}{@{\hspace{1pt}}c@{\hspace{1pt}}c@{\hspace{1pt}}c@{\hspace{1pt}}c@{\hspace{1pt}}}
         \includegraphics[width=0.24\columnwidth]{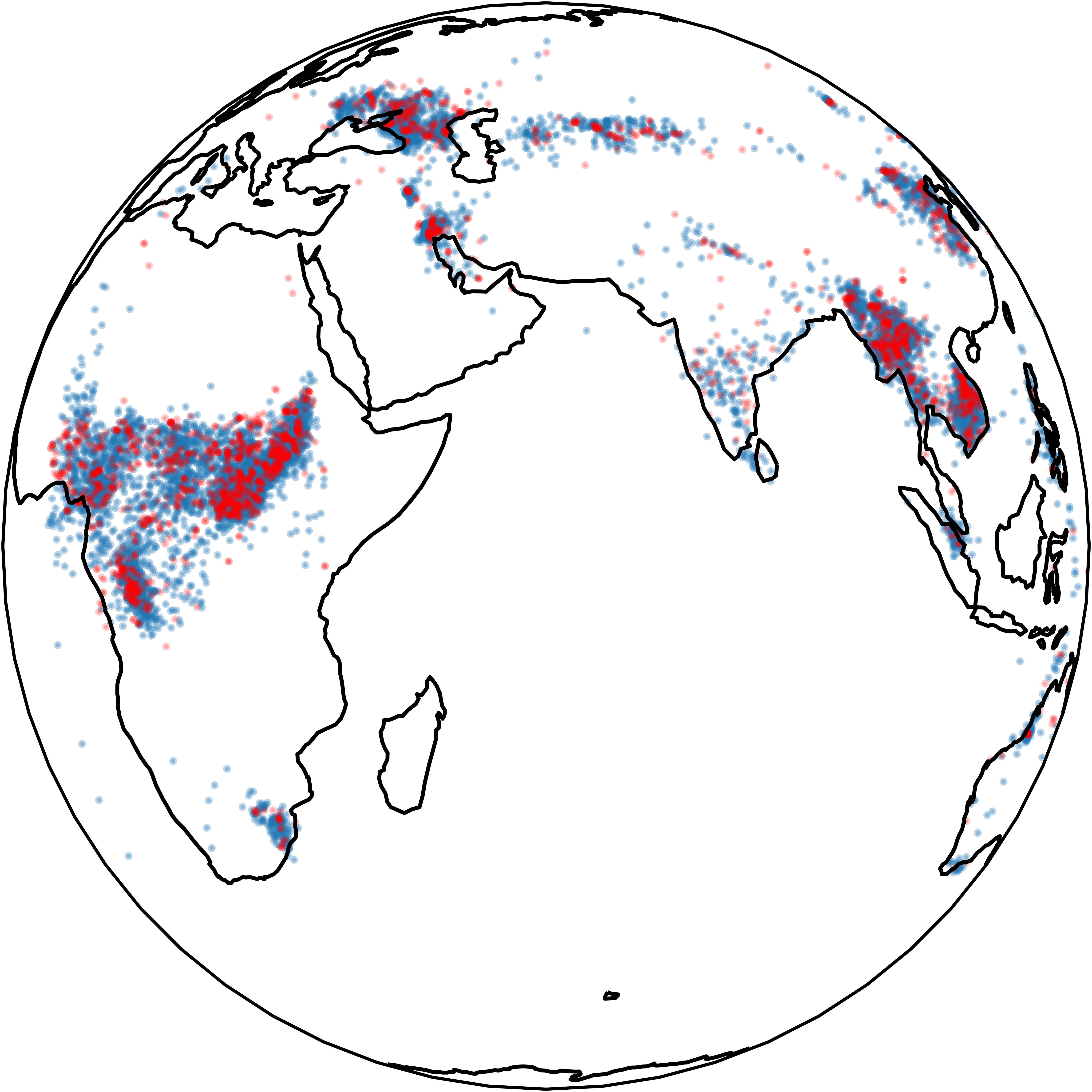}&
         \includegraphics[width=0.24\columnwidth]{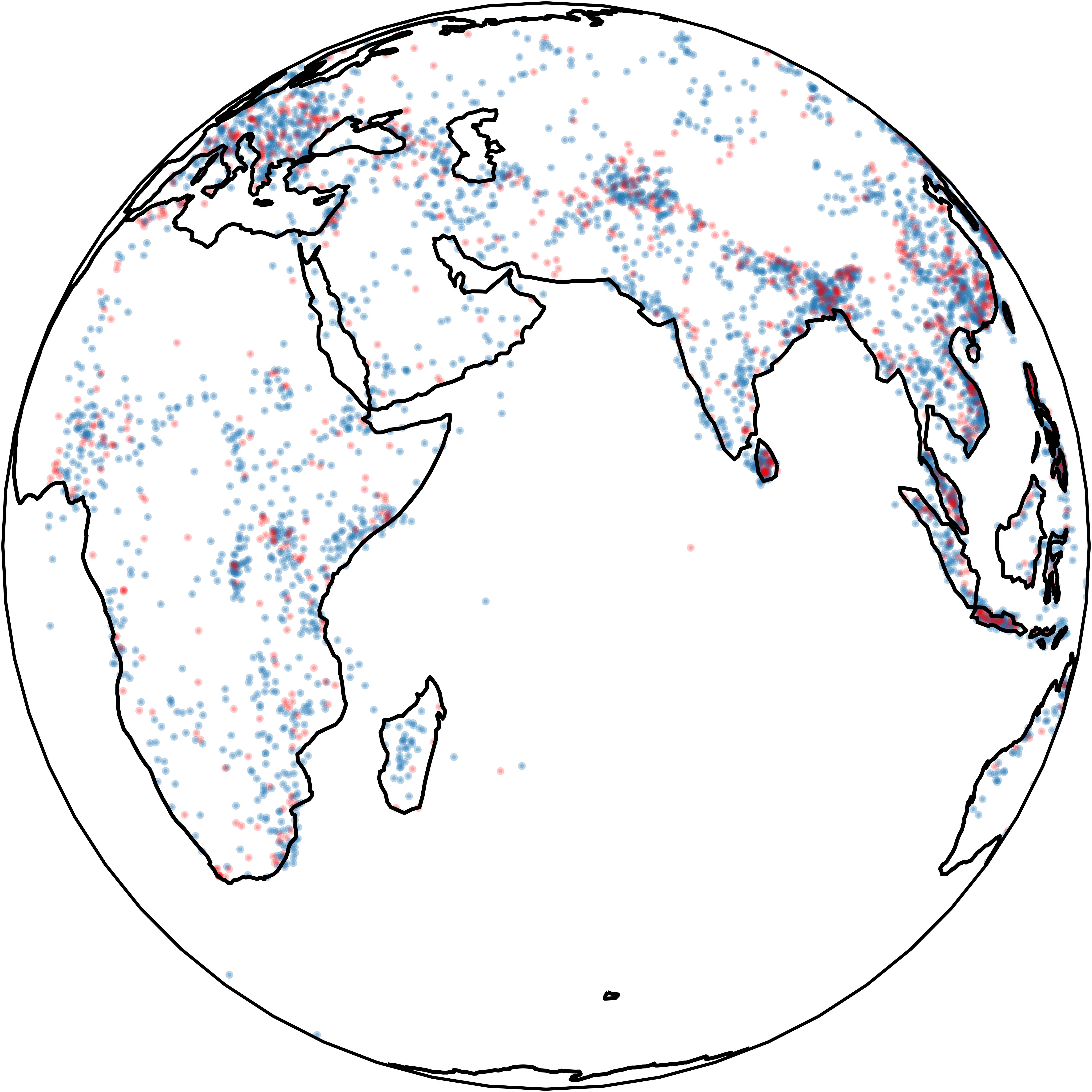}&
         \includegraphics[width=0.24\columnwidth]{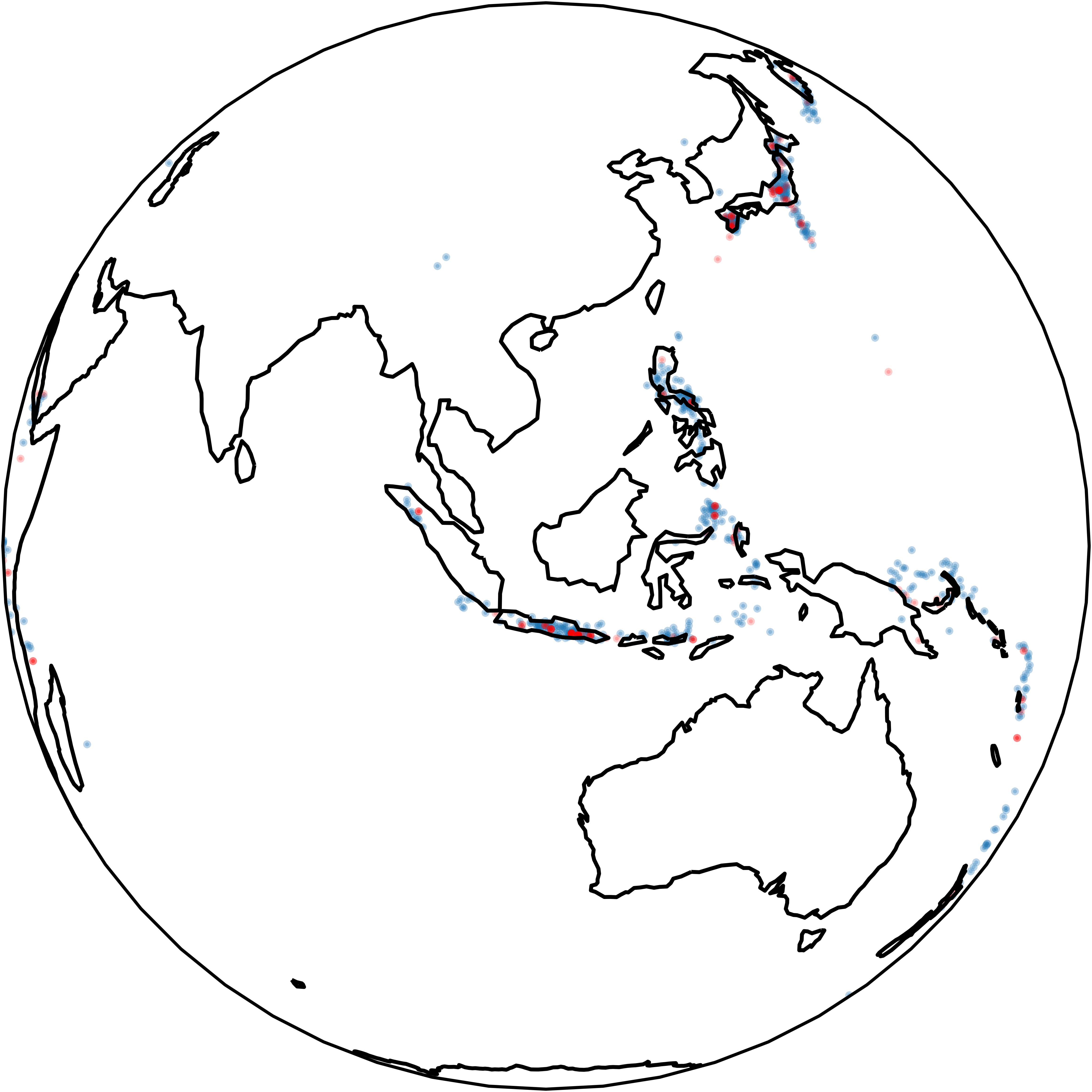}&
         \includegraphics[width=0.24\columnwidth]{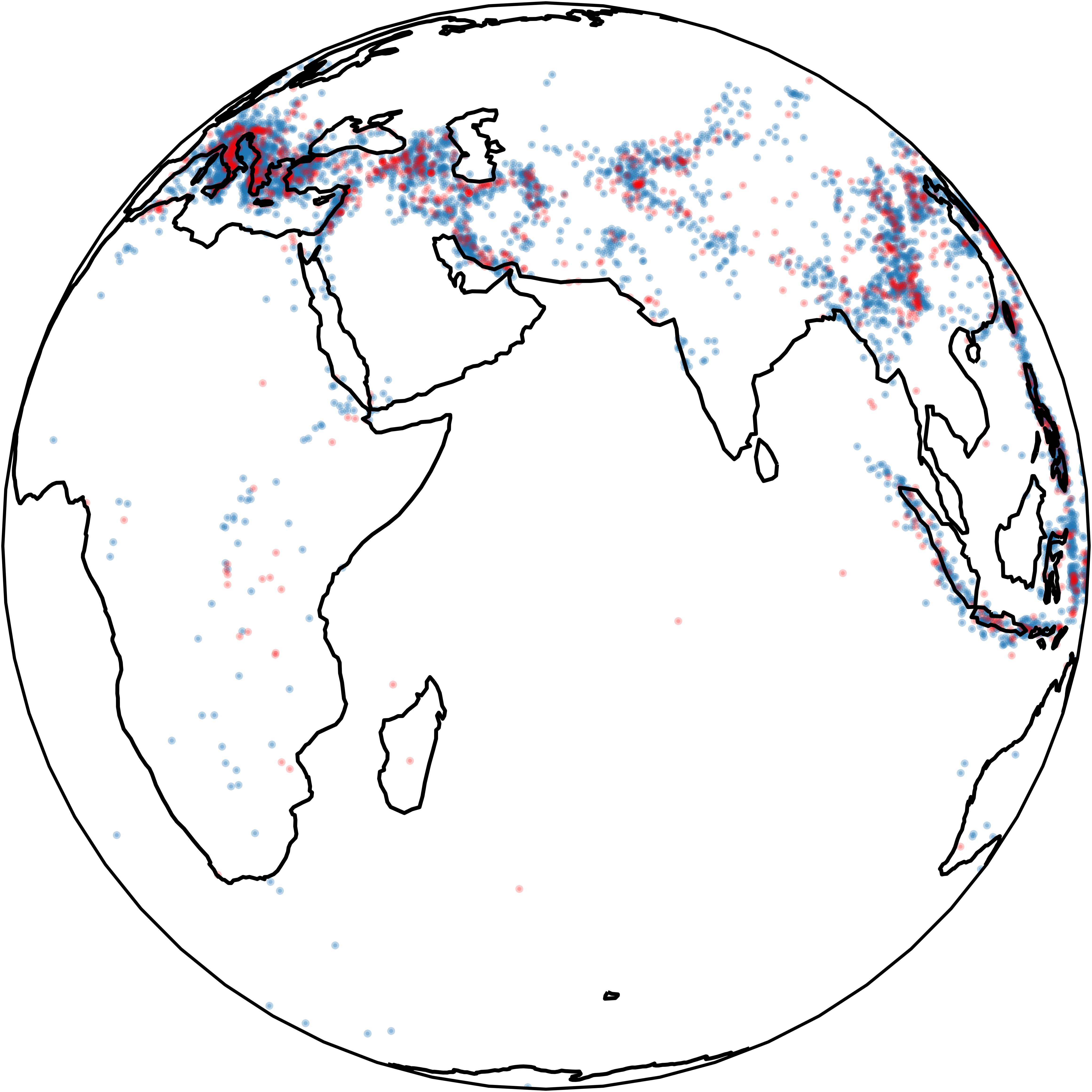}\\ {\scriptsize Fire} & {\scriptsize Flood} &  {\scriptsize Volcano} & {\scriptsize Quakes}
    \end{tabular}
    \caption{Earth and Climate dataset: generated samples from the trained CNFM in blue, test samples in red. See table \ref{tab:earth_results} for quantitative results.  }
    \label{fig:earth_qualitative}
\end{figure}

In this experiment we considered the Earth and Climate dataset curated in \cite{mathieu2020riemannian}. This dataset contains locations of earthquakes, floods, fires, and volcano eruptions on earth, represented as point locations on the 2D sphere, $\gS^2\subset \Real^3$. The target path $p$ is defined as in \eqref{e:path_sphere} with $\kappa_1=55$K (best out of $\kappa_1\in \set{5\text{K},55\text{K},500\text{K}}$). We used the same architecture used in \cite{rozen2021moser}, a MLP with $6$ layers of $512$ neurons, PPD order $\ell=2$. Table \ref{tab:earth_results} depicts the negative log likelihoods (NLLs) scores, where CNFM improves state of the art by a large margin, where the runner-up is Moser Flow \cite{rozen2021moser}. Riemannian CNF and other baselines are taken from \cite{mathieu2020riemannian}. Figure \ref{fig:earth_qualitative} visualizes generated samples (blue) and test data samples (red). 

\begin{table}
\centering
\resizebox{\columnwidth}{!}{%
\begin{tabular}{ |c||c|c|c|c|} 
 \hline
 Dataset & Earthquake & Flood & Fire & Volcano \\ 
 \hline \hline
 Mixture vMF & $0.59{\scriptstyle\pm0.01}$ &  $1.09{\scriptstyle\pm0.01}$& $-0.23{\scriptstyle\pm0.02} $ & $-0.31{\scriptstyle\pm0.07} $\\ 
 Stereographic & $0.43{\scriptstyle\pm0.04}$ &  $0.99{\scriptstyle\pm0.04}$& $-0.40{\scriptstyle\pm0.06} $& $-0.64{\scriptstyle\pm0.20} $\\ 
 Riemannian & $0.19{\scriptstyle\pm0.04}$ &  $0.90{\scriptstyle\pm0.03}$& $-0.66{\scriptstyle\pm0.05} $& $-0.97{\scriptstyle\pm0.15} $\\ 
 Moser Flow & $-0.09{\scriptstyle\pm0.02}$ &  $0.62{\scriptstyle\pm0.04}$& $-1.03{\scriptstyle\pm0.03} $& $-2.02{\scriptstyle\pm0.42} $\\ 
 CNFM & $\mathbf{-0.38}{\scriptstyle\pm0.01}$ &  $\mathbf{0.25}{\scriptstyle\pm0.02}$& $\mathbf{-1.40}{\scriptstyle\pm0.02} $& $\mathbf{-2.38}{\scriptstyle\pm0.16} $\\ 
 \hline
\end{tabular} }
\caption{Negative log likelihood scores on the Earth and Climate Dataset \cite{mathieu2020riemannian}.}
\label{tab:earth_results}
\end{table}

\begin{figure}
    \centering
    \includegraphics[width=\columnwidth]{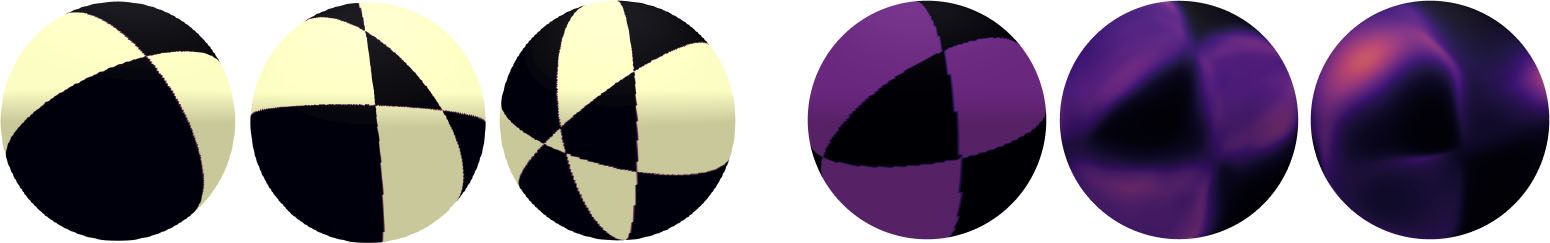}
    \caption{Left triplet shows the densities $r_k$ for $k=2,3,4$ on random cuts $\gS^2\subset \gS^{15}$; right triplet visualizes the case $k=3$ (on a different random cut) from Table \ref{tab:nll_spheres} with CNFM model density in the middle, and S-FFJORD density on the right.}
    \label{fig:cuts}
\end{figure}
\subsection{Higher dimensional spheres}




In this experiment we test the scaling of CNFM to higher dimensional manifold data. We construct a family of challenging probability distributions, denoted $r_k$, on $\gS^{15}$ and compare CNFM to several baselines. We start by defining $r_k$ over $\gS^{15}\subset \Real^{16}$: Henceforth, denote $d=15$, and consider an orthogonal set $v_1,\ldots,v_k$, where $1\leq k \leq d+1$. Let $s(x)=\prod_{i=1}^{k} \sign(x^T v_i) $. Define the probability density:
\begin{equation}
    r_k(x) = \frac{2}{\abs{\gS^d}}\begin{cases}
    1 & \text{if } s(x)=1\\
    0 & \text{if } s(x)=-1
    \end{cases}
\end{equation}
To see $r_k$ is indeed a probability density, note that the transformation $x=(x_1,\ldots,x_{d+1})\mapsto (-x_1,\ldots,x_{d+1})$ is a volume preserving transformation of $\gS^d$ and maps the set $\Omega_+=\gS^d\cap\set{x\in\Real^{d+1}\vert s(x)=1}$ to $\Omega_-=\gS^d\cap\set{x\in\Real^{d+1}\vert s(x)=-1}$, and vise versa. This means that $\int_{\Omega_+} dV_x = \int_{\Omega_-} dV_x$ and since $\gS^d=\Omega_+\cup \Omega_-$ we have that $\int_{\Omega_+} dV_x= \abs{\gS^d}/2$. Generating samples from $r_k$ can be done by randomizing a uniform sample $x$ over $\gS^d$, if $s(x)=1$, keep $x$, otherwise take $(-x_1,x_2,\ldots,x_{d+1})$. Figure \ref{fig:cuts}-left depicts several examples of this density by visualizing random $\gS^2$ cuts in $\gS^{15}$; as $k$ increases the complexity of density increases. We created datasets for $k=2,3,4$ with $45$K train samples and $5$K test samples. 
\begin{wraptable}[8]{r}{0.4\columnwidth}
\centering
\resizebox{0.4\columnwidth}{!}{%
\begin{tabular}{|@{\hspace{3pt}}c@{\hspace{3pt}}|@{\hspace{3pt}}c@{\hspace{3pt}}c@{\hspace{3pt}}c@{\hspace{3pt}}|}
\hline
            & 2       & 3      & 4      \\ \hline \hline
vMF-MM & $1.23$   & $1.31$  & $1.33$ \\
S-FFJORD    & $0.77$   & $0.97$  & $1.04$ \\
CNFM        & $\mathbf{0.73}$   & $\mathbf{0.83}$  & $\mathbf{0.95}$  \\ \hline    
\end{tabular}} \vspace{-0pt}
\caption{NLLs on $\gS^{15}$.  }
\label{tab:nll_spheres}
\end{wraptable}
For baselines we use: vMF mixture models (vMF-MM) with $1$K and $10$K centers randomized from the training data, and scaling $\kappa$ was chosen to be the optimal for the test set. This was done to compare to the best possible vMF-MM model.  Furthremore, we compared to a version of manifold CNF \cite{lou2020neural,mathieu2020riemannian,falorsi2020Neural}: We consider the stereographic projection of the sphere $\Psi:\Real^d \too \gS^d $, and used FFJORD~\cite{grathwohl2018ffjord} code adapted to the spherical case, denoted as S-FFJORD.  In this baseline, computing log probabilities over the sphere is done by  correcting for the stereographic projection, $\log p(\Psi(u)) = \log p(u) - \frac{1}{2}\log \det (D_\Psi(u)^T D_\Psi(u))$, where $u\in\Real^d$, $\log p(u)$ is the Euclidean log probability learned by FFJORD,  $D_\Psi(u)\in\Real^{(d+1)\times d}$ is the matrix of partials of $\Psi$. Table \ref{tab:nll_spheres} reports the NLL scores of CNFM and the baselines across this dataset. Figure \ref{fig:cuts}-right depicts an example of random $\gS^2$ cut of $\gS^{15}$ for the $k=3$ case.

\begin{wrapfigure}[9]{r}{0.5\columnwidth}
    \includegraphics[width=0.53\columnwidth]{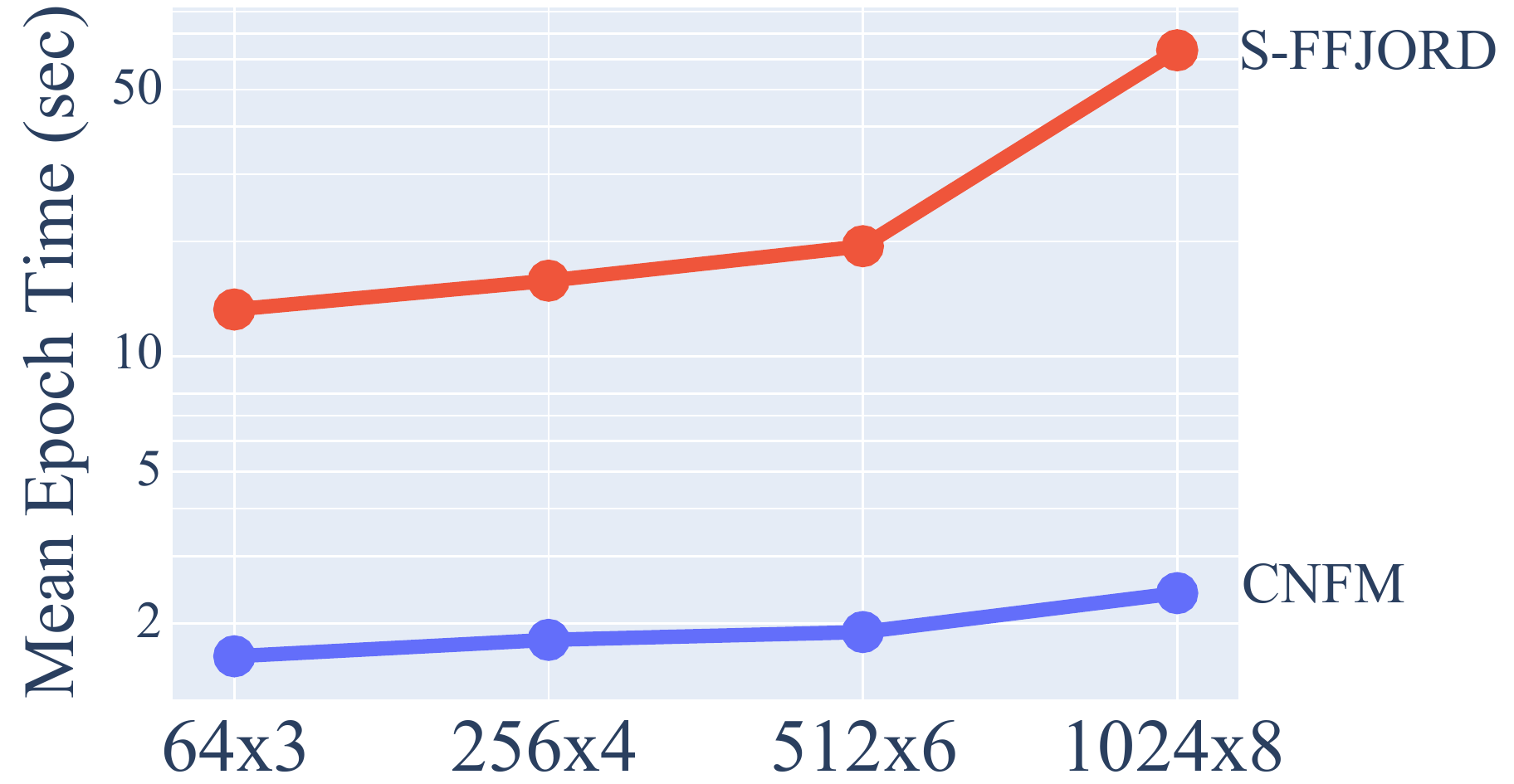}\vspace{-5pt}
  \caption{Timings.}\label{fig:times_and_plot}
\end{wrapfigure}
The above results are reported using an MLP with 3 layers of 64 neurons for CNFM and an equivalent architecture of S-FFJORD, with both methods running for about $4$K seconds. 
In Figure \ref{fig:times_and_plot} we compared typical epoch running times for this and larger architecture types for CNFM and FFJORD training. Note that the time difference (in log scale) further increases for larger architectures. 



\begin{figure}[t]
    \centering
    \includegraphics[width=\hsize]{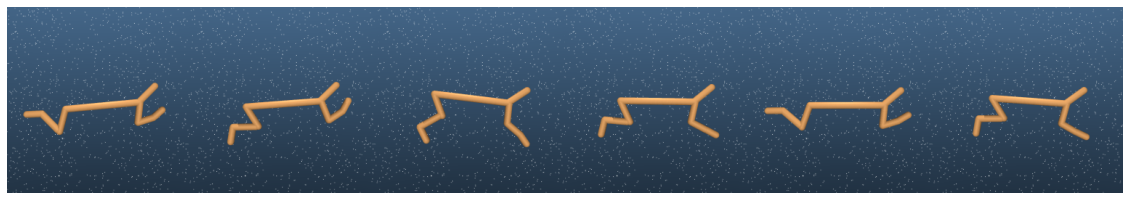}
    \includegraphics[width=\hsize]{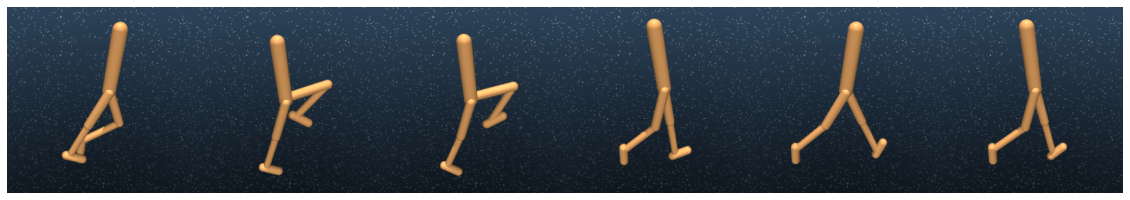}
    \includegraphics[width=\hsize]{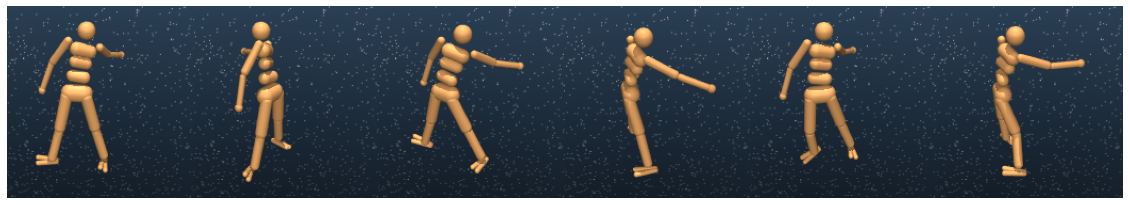}
    \caption{Uncurated samples computed with the trained CNFM on product manifolds representing the robot's state space: Cheetah (top), Walker (middle), and Humanoid (bottom). }
    \label{fig:robots_samples}
\end{figure}
\subsection{Product of manifolds - Robotics}
\begin{figure}[t]
    \centering
    \includegraphics[width=\hsize]{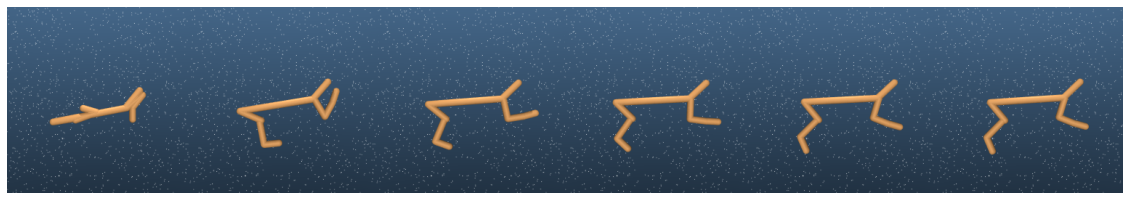}
    \includegraphics[width=\hsize]{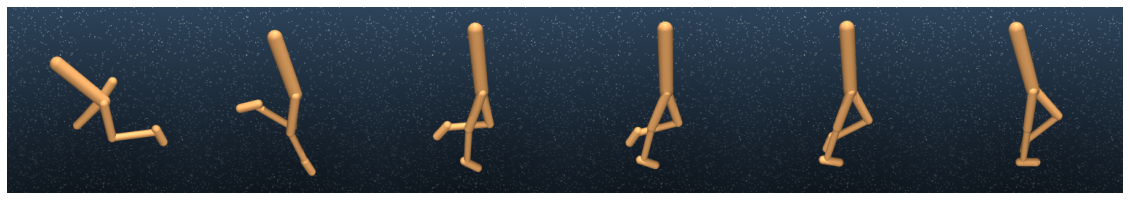}
    \includegraphics[width=\hsize]{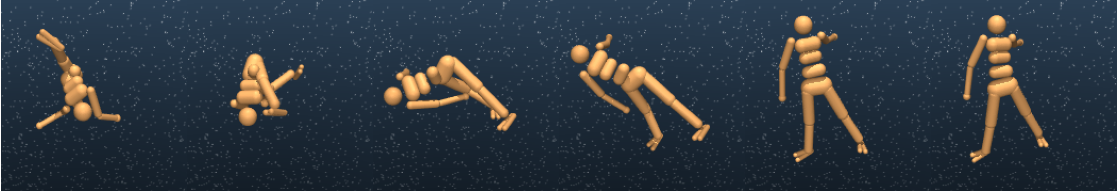}
    \caption{Noise to data paths computed with the trained CNFM on product manifolds representing robot's state space: Cheetah (top), Walker (middle), and Humanoid (bottom).}
    \label{fig:robots_noise_to_data}
\end{figure}

In the last experiment we worked with robotics data generated with the physics and reinforcement learning engine MuJoCo~\cite{tassa2020dmcontrol}. For each of the three robot types, Walker (2D), Cheetah (2D), and Humanoid (3D), we randomized $17.5$K samples from $50$ simulated trajectories consisting of $500$ observation each. The state space for the 2D robots is modeled as the product manifold $\gM= \Real^{3}\times  (\gS^1)^6$, where $\Real^3$ represents position, and $\gS^1$ represents 2D rotations of a single joint. The state space for the 3D robot is $\gM=\Real^3\times(\gS^1)^8\times(\gS^3)^6$, where $\gS^3$ represents 3D rotations of a joint (via quaternions). We used the target path $p$ on the product manifold as described in Section \ref{ss:target_path}. For each robot type, Figure \ref{fig:robots_samples} depicts uncurated samples from the trained CNFM, and Figure \ref{fig:robots_noise_to_data} shows a path of noise to data, \ie, $\phi_t(x)$, $t\in[0,1]$, where $x\sim p_0$. 
The generated samples are qualitatively similarly to data samples. More examples are in Appendix \ref{a:more_experiments}.

\section{Limitations and Future Work}
Scaling CNFM to even higher dimensions (\ie, $d>100$), \eg, for image data, requires some more work. We identify two main challenges: (i) using \texttt{logsumexp} and stochastic approximation for $\mathrm{div}(v)$ introduces a non-trivial approximation error (and bias) to the gradient estimation of the PPD; and (ii) the PPD loss has very different scales for different values of $t$, which entails conditioning.  
\begin{wrapfigure}[8]{r}{0.3\columnwidth}\hspace{-6pt}
    \includegraphics[width=0.3\columnwidth]{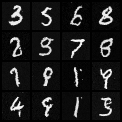}
\end{wrapfigure}
Using CNFM as-is on the MNIST dataset ($d=784$) with a standard batch size of 128 results in samples shown in inset. Although generation quality does not match SOTA CNF models, the training process remains stable despite the high dimensional biased gradient estimation of the loss. We leave scaling CNFM to images to future work.

\section{Conclusions}
We have introduced CNFM, a framework for matching a target density path and the density path generated by a CNF. The CNFM is based on minimizing a novel Probability Path Divergence (PPD) that does not require sampling of model densities and therefore is easier to train and to apply to manifolds. The PPD is shown to upper bound standard divergences, and can work with a rather flexible family of target paths on manifolds. Empirically, CNFM was shown to facilitate CNF training, scaling for the first time to manifolds of moderate dimension, improving training time, and producing state of the art samplings and log likelihoods.

\section{Acknowledgments}
HB was supported by a grant from Israel CHE Program for Data Science Research Centers and Carolito Stiftung (WAIC). SC was supported by the
Engineering and Physical Sciences Research Council
(grant number EP/S021566/1).

%

%


\bibliography{CCNF}
\bibliographystyle{icml2022}

\newpage
\appendix
\onecolumn


\section{Proof of Theorem \ref{thm:lmc}}\label{a:proof_lmc}
Since $v\in \mathfrak{X}(\gM)$ it is locally Lipschitz. Since $v$ is bounded it satisfies in particular $$\int_0^1 \int_\gM \abs{v(t,x)}p_t(x)\, dV_x dt < +\infty.$$
Therefore, according to the Mass Conservation Formula Theorem (see \eg, \cite{villani2009optimal}) \eqref{e:LMC_phi_generates_alpha_t}
 holds iff 
\begin{equation}\label{ea:mass_conserve}
\partial_t p_t + \mathrm{div}(p_t v) = 0,
\end{equation}
where $\mathrm{div}$ denotes the divergence operator over the manifold $\gM$. We assumed $p_t>0$ and therefore dividing both sides with $p_t$ leads to 
$$\frac{\partial_t p_t}{p_t}+\frac{\ip{\nabla_x p_t, v}+p_t \mathrm{div}(v)}{p_t}=0.$$
where we used that $\mathrm{div}(fv)=\ip{\nabla_x f, v} + f\mathrm{div}(v)$ for  $f\in\mathfrak{P}(\gM)$ and $v\in \mathfrak{X}(\gM)$. Finally noting that $\partial_t \log p_t= \frac{\partial_t p_t}{p_t}$, and $\nabla_x \log p_t= \frac{\nabla_x p_t}{p_t}$ we get that \eqref{e:lmc} is equivalent to  \eqref{ea:mass_conserve}. $\qed$

\section{Proof of Lemma \ref{lem:d_form}}\label{a:proof_d_form}
Given a time dependent vector field $v\in\mathfrak{X}(\gM)$, a diffeomorphism two parameter family $\Phi_{t,t_0}:\gM\too\gM$ can be defined via the following Ordinary Differential Equation (ODE):
\begin{align}\label{e:Phi}
\begin{cases}
\frac{d}{dt}\Phi_{t,t_0}(x) = v(t,\Phi_{t,t_0}(x)) & \\
\Phi_{t_0,t_0}(x) = x & 
\end{cases}
\end{align}
The CNF diffeomorphism is defined by $\phi_t=\Phi_{t,0}$. Now, consider a smooth function $u(t,x)$, $u:[0,1]\times \gM\too \Real$, then
\begin{equation}\label{e:u}
    \partial_t\vert_{t=t_0}\brac{ u(t,\Phi_{t,t_0}(x))} = \partial_t u(t_0, x) + \ip{\nabla_x u(t_0,x),v(t_0,x)}.
\end{equation}
From Theorem \ref{thm:lmc} we have that 
\begin{equation*}
    \partial_t \log q_t + \ip{\nabla_x \log q_t , v} + \mathrm{div}(v)=0
\end{equation*}
for all $t\in [0,1]$ and $x\in \gM$. Subtracting that in our loss we get 
\begin{align}\label{e:proof_d_form_step}
    \dist_\ell(p \bb q) &= \E_{t,x\sim p_t} \abs{\partial_t \log\frac{p_t}{q_t} + \ip{\nabla_x\log \frac{p_t}{ q_t} , v }}^\ell
\end{align}
Now using \eqref{e:u} with $u(t,x)=\log\frac{p_t(x)}{q_t(x)}$, we get
\begin{align*}
    &\partial_t\vert_{t=t_0} \log\frac{ p_t(\Phi_{t,t_0}(x))}{ q_t(\Phi_{t,t_0}(x))}=\\&\partial_t\log\frac{ p_{t_0}(x)}{ q_{t_0}(x)}+\ip{\nabla_x \log\frac{ p_{t_0}(x)}{ q_{t_0}(x)} , v(t_0,x)}
\end{align*}
Plugging this in \eqref{e:proof_d_form_step} with $t=s$ and $t_0=t$ we get:
\begin{align*}
    \dist_\ell(p \bb q) &= \E_{t,x\sim p_t} \abs{ \partial_s\vert_{s=t} \log\frac{ p_s(\Phi_{s,t}(x))}{ q_s(\Phi_{s,t}(x)) } }^\ell \\
    &= \E_{t,x\sim q_t} \frac{p_t(x)}{q_t(x)} \abs{ \partial_s\vert_{s=t} \log\frac{ p_s(\Phi_{s,t}(x))}{ q_s(\Phi_{s,t}(x)) } }^\ell \\
    &= \E_{t,x\sim p_0} \frac{p_t(\phi_t(x))}{q_t(\phi_t(x))}\abs{ \partial_s\vert_{s=t} \log\frac{ p_s(\Phi_{s,t}(\phi_t(x)))}{ q_s(\Phi_{s,t}(\phi_t(x))) } }^\ell
    \\
    &= \E_{t,x\sim p_0} \frac{p_t(\phi_t(x))}{q_t(\phi_t(x))} \abs{ \partial_s\vert_{s=t} \log\frac{ p_s(\phi_{s}(x))}{ q_s(\phi_{s}(x)) } }^\ell \\
    &= \E_{x\sim p_0 }\int_0^1  \frac{p_t(\phi_t(x))}{q_t(\phi_t(x))} \abs{ \partial_t \log\frac{ p_t(\phi_{t}(x))}{ q_t(\phi_{t}(x)) } }^\ell dt
\end{align*}
where in the third equality we used the fact that $\phi_t(x)\sim q_t$ if $x\sim p_0$; in the fourth equality we used the fact that $\Phi_{s,t}(\phi_t(x))=\Phi_{s,t}(\Phi_{t,0}(x))=\Phi_{s,0}(x)=\phi_s(x)$. \qed

\section{Additional details for the proof of Theorem \ref{thm:relation_to_divergence}}
\label{a:relation_to_divergence}
We add here details of the proof of Theorem \ref{thm:relation_to_divergence} missing from the main paper. 

First, we prove that for any $t>0$, $ \ell(1-t^{\frac{1}{\ell}})\nearrow -\log(t)$, that is monotonically increasing and converging to $-\log (t)$ as $\ell\too\infty$. Fix $t>0$, and define the function $$f(s)=\frac{(1-t^{s})}{s}$$ where $s\in(0,1)$. Now, using L'H\^{o}pital's rule:
\begin{equation*}
 \lim_{s\shortdownarrow 0} f(s)= \lim_{s\shortdownarrow 0}\frac{-\log(t)t^{s}}{1}=-\log(t)   
\end{equation*}
Therefore in particular $\lim_{\ell\too\infty} \ell(1-t^{\frac{1}{\ell}})=-\log(t)$. Monotonicity follows from the fact that for all $s\in(0,1)$ and $t>0$
\begin{equation*}
    f'(s) = \frac{-\log(t)st^s+t^s-1}{s^2} = \frac{t^s}{s}\parr{\frac{1-t^{-s}}{s}-\log(t)}\leq0
\end{equation*}
The inequality can be justified by first noting that $t^s/s >0$. Second, let $0<t=\exp(r)$ we get that
\begin{align*}
    \frac{1-t^{-s}}{s}-\log(t) \leq 0
\end{align*}
which is true iff
\begin{align*}
    \frac{1-\exp(-rs)}{s}-r \leq 0
\end{align*}
which is true iff 
\begin{align*}
    1-rs \leq \exp(-rs) 
\end{align*}
which is true iff for all $u\in \Real$
\begin{align*}
    1-u \leq \exp(-u) 
\end{align*}
which is true since $1-u$ is tangent to $\exp(-u)$ at $u=0$, and $\exp(u)$ is convex. Since $f'(s)$ is monotonically decreasing in $s$, $\ell(1-t^{\frac{1}{\ell}})$ is increasing as $\ell\too\infty$. 

We are now ready to justify \eqref{e:lim_switch}. First let
\begin{equation*}
    f_\ell(x)=\ell \parr{1-\brac{\frac{p_T(x)}{q_T(x)}}^{\frac{1}{\ell}}}
\end{equation*} 
We showed that $f_\ell(x)\nearrow f(x)=-\log\frac{p_T(x)}{q_T(x)}$. Furthermore, $f_\ell$ are all integrable since
\begin{align*}
&\int_\gM \ell\abs{1-\brac{\frac{p_T(x)}{q_T(x)}}^{\frac{1}{\ell}}}q_T(x)dV_x  \\
&\leq \ell\int_\gM  q_T(x)dV_x +   \ell \int_\gM p_T(x)^{\frac{1}{\ell}}q_T^{1-\frac{1}{\ell}} dV_x \\
&\leq \ell + \ell\brac{\int_\gM p_T(x)dV_x}^{\frac{1}{\ell}}\brac{\int_\gM q_T(x)dV_x}^{1-\frac{1}{\ell}} \\ &= 2\ell
\end{align*}
Where in the first inequality we used the triangle inequality, and in the second inequality we used Holder inequality with $\frac{1}{\ell}+\frac{\ell-1}{\ell} = 1$. 

We assume that $$D_f(p_T \bb q_T)=\int_\gM f(x) q_T(x)dV_x<\infty.$$ Since $f_\ell(x)\leq f(x)$ and both $f_\ell,f$ are integrable we have that 
$$\int_\gM f_\ell(x)q_T(x)dV_x \leq \int_\gM f(x) q_T(x)dV_x < \infty$$ for all $\ell$. Therefore, the Monotone Convergence Theorem (see Theorem 2.8.2 in \cite{bogachev2007measure}) implies that 
\begin{equation*}
    \lim_{\ell \too \infty} \int_\gM f_\ell(x) q_T(x)dV_x = \int_\gM f(x) q_T(x)dV_x
\end{equation*}
Namely,
\begin{equation*}
    \lim_{\ell \too \infty} \E_{x\sim q_T} \ell \parr{1-\brac{\frac{p_T(x)}{q_T(x)}}^{\frac{1}{\ell}}}  = -\E_{x\sim q_T} \log\frac{p_T(x)}{q_T(x)} 
\end{equation*}

\section{Numerically stable derivative of the normalizing constant of the vMF} \label{a:constant_of_vmf}
The log of the normalizing constant of the vMF has the form
\begin{equation*}
    \log C_p(\kappa) = \parr{\frac{p}{2}-1} \log\kappa - \frac{p}{2}\log(2\pi) - \brac{\kappa + \log \texttt{ive}(\frac{p}{2}-1,\kappa)}
\end{equation*}
where 
$\texttt{ive}(\nu,\kappa) = \texttt{iv}(\nu,\kappa)\exp(-\kappa)$. 
Now, the $ \log \texttt{ive}(\frac{p}{2}-1,\kappa)$ is stable but its derivative is not. Therefore we will define a new function and its derivative: $\texttt{logive}(\nu,\kappa)$. Its forward will be defined by;
\begin{equation*}
    \texttt{logive}(\nu,\kappa) = \log \parr{\texttt{ive}(\nu,\kappa)},
\end{equation*}
and for its derivative we first note:
\begin{align*}
    \partial_\kappa \log \texttt{ive}(\nu,\kappa) &= \frac{\partial_\kappa  \texttt{ive}(\nu,\kappa)}{ \texttt{ive}(\nu,\kappa)}=\frac{\texttt{ive}(\nu-1,\kappa) - \texttt{ive}(\nu,\kappa)\brac{\frac{\nu+\kappa}{\kappa}}}{\texttt{ive}(\nu,\kappa)} \\ &= \frac{\texttt{ive}(\nu-1,\kappa) }{\texttt{ive}(\nu,\kappa)}-\brac{\frac{\nu+\kappa}{\kappa}}=\frac{\texttt{ive}(\nu-1,\kappa) }{\texttt{ive}(\nu,\kappa)} - \frac{\nu}{\kappa}-1
\end{align*}
For high dimensions the $\texttt{ive}$ ratio is numerically unstable and several approximations have been suggested. In particular \cite{ruiz2016new} suggest the following lower and upper bounds:  \begin{align*}
     \frac{\nu-\frac{1}{2}+\sqrt{(\nu+\frac{1}{2})^2+\kappa^2}}{\kappa} > \frac{\texttt{ive}(\nu-1,\kappa)}{\texttt{ive}(\nu,\kappa)} > \frac{\nu-1+\sqrt{(\nu+1)^2+\kappa^2}}{\kappa}
\end{align*}
Similar to \cite{oh2019radial} we take the average of the higher and lower bound (see \cite{oh2019radial} for empirically demonstrating the quality of this approximation):
\begin{align*}
    \partial_\kappa \log \texttt{ive}(\nu,\kappa) =\frac{\texttt{ive}(\nu-1,\kappa) }{\texttt{ive}(\nu,\kappa)} - \frac{\nu}{\kappa}-1 \approx \frac{-1.5+\sqrt{(\nu+1)^2+\kappa^2}+\sqrt{(\nu+\frac{1}{1})^2+\kappa^2}}{2\kappa}-1
\end{align*}
and this is defined as the derivative of $\texttt{logive}$. 
\section{Vector field representation} \label{a:vector_field_rep}
We describe how we represent the parametric part of our system, namely the vector field $v_\theta \in  \mathfrak{P}(\gM)$, for the different manifold types we consider: Euclidean space, spheres and product manifolds.  In the \emph{Euclidean} case we use an MLP $v_\theta:\Real^{d+1}\too\Real^d$, where $d$ is the dimension of $\gM$. We use standard Euclidean inner product, gradient $\nabla$, and divergence $\mathrm{div} = \nabla \cdot$ for computing the PPD (\eqref{e:d}), where the path $p$ defined in \eqref{e:path_euc}. 
In the \emph{sphere} case $\gS^{d}\subset\Real^{d+1}$, similarly to \cite{rozen2021moser} we define $v$ to be constant in the normal direction to the sphere and produce tangent vectors via the tangent projection operator
\begin{equation}\label{e:v_model}
  v_\theta(t,x) = \bigg ( \mI - \frac{xx^T}{\norm{x}^2} \bigg )w_\theta\parr{t,\frac{x}{\norm{x}}},  
\end{equation}
where $w_\theta: \Real^{d+2}\too \Real^{d+1}$ is an MLP. The inner product on the sphere is the induced Euclidean one, \ie, for $v,u\in T_x \gS^d$ we have $\ip{u,v}=u^Tv$. For $v_\theta$ defined in \eqref{e:v_model}, the Riemannian gradient and divergence coincide with the Euclidean gradient $\nabla$ and divergence $\nabla\cdot$. $p$ is defined as in \eqref{e:path_sphere}. 
For notational simplicity we explain the \emph{product manifold} implementation for two manifolds $\gM=\Real^{d_1}\times \gS^{d_2}$, where the extension to product of $N$ manifolds is similar. The tangent vector field is a function of the form $v_\theta:\Real^{d_1+1}\times\Real^{d_2+2}\too \Real^{d_1}\times\Real^{d_2+1}$. For $(t_1,x_1,t_2,x_2)\in \Real^{d_1+1}\times\Real^{d_2+2}$ we let 
\begin{equation*}
    v_\theta(t_1,x_1,t_2,x_2) = \begin{bmatrix}
     v_1\bigg(t_1,x_1,t_2,\frac{x_2}{\norm{x_2}}\bigg)  \\  \bigg(\mI - \frac{x_2x_2^T}{\norm{x_2}^2}\bigg)  v_2\bigg(t_1,x_1,t_2,\frac{x_2}{\norm{x_2}}\bigg) 
    \end{bmatrix}
\end{equation*}
where $v_1,v_2$ are MLPs.  The inner product $(v_1,v_2),(u_1,u_2)\in T_x \gM$ is defined by $\ip{v_1,u_1}+\ip{v_2,u_2}$; the gradient as $\nabla  = (\nabla_1 , \nabla_2)^T$, where $\nabla_1$ is the Euclidean gradient w.r.t.~$x_1$, and $\nabla_2$ is the Euclidean gradient w.r.t.~$x_2$; the divergence $\mathrm{div}(v) = \nabla_1 \cdot v_1 + \nabla_2 \cdot v_2$. Lastly, $p$ is defined as in \eqref{e:path} with kernel \eqref{e:product_kernel}. 

\section{Experimental Details}
\subsection{Toy densities on $\Real^2$ and $\gS^2$}
For the $\Real^2$ datasets we used a 3 layer MLP with hidden dimension 256. We trained with Adam optimizer with learning rate $1e-4$, batch size $1000$, $\sigma_1=0.01$ and $\ell=1$. The searched parameters across learning rates are $\{1e-3, 5e-4, 1e-4 \}$ and $\sigma_1\in\{0.005, 0.01, 0.05\}$.
For the $\gS^2$ datasets we used a 6 layer MLP with hidden dimension 512. We trained with Adam optimizer with learning rate $1e-4$, batch size $1000$, $\kappa=5000$ and $\ell=1$. 

\subsection{Earth and climate datasets}
For the earth and climate datasets we used a 6 layer MLP with hidden dimension 512. We trained with Adam optimizer with learning rate $1e-4$, batch size $1000$, $\kappa=55K$ and $\ell=2$. The searched parameters across $\kappa$ are $\{5K, 55K, 500k \}$.

\subsection{Higher dimensional spheres}
We ran experiments on $\gS^{15}$ for different $k=2,3,4$ values. The architecture used was a 3 layer MLP with hidden dimension 64, Adam optimizer with learning rate $1e-3$, batch size $7000$, $\kappa=5K$ and $\ell=2$.
We searched over learning rates $\{ 1e-3,1e-4,1e-5\}$.

The S-FFJORD baseline is as described in the paper. We used the architecture used for the 2D toy experiments in the FFJORD paper, as published in the official FFJORD code repository. 
We run both CNFM and S-FFJORD with approximate divergence computation using the Hutchinson estimator.
\subsection{Product of manifolds - Robotics}
For the robotics datasets we used a 6 layer MLP with hidden dimension 512. We trained with Adam optimizer with learning rate $1e-4$, batch size $1000$, $\kappa=55K$ and $\ell=1$. The searched parameters across $\kappa$ are $\{5K, 55K\}$.

\section{Extra Experimental Results}\label{a:more_experiments}
\begin{figure}
    \centering
    \includegraphics[width=0.49\hsize]{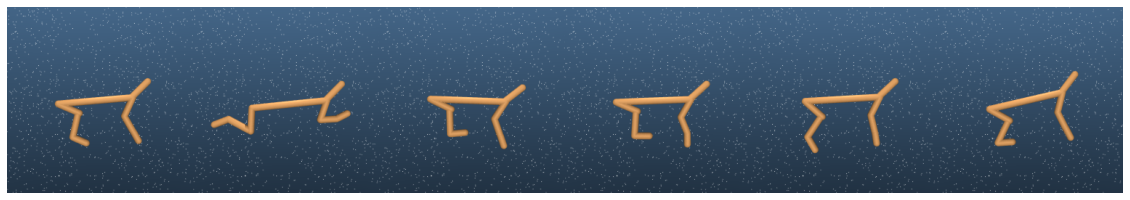}\hspace{-2.2mm}
    \includegraphics[width=0.49\hsize]{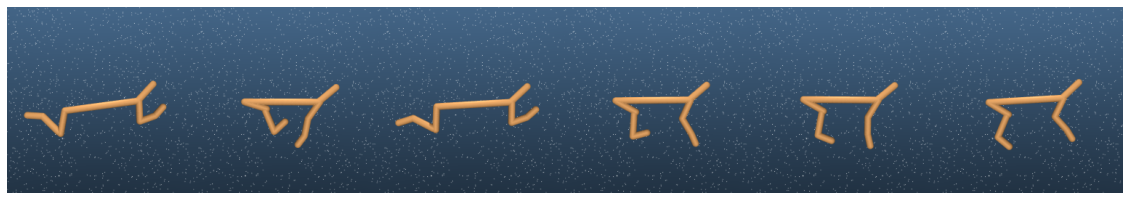}
     \includegraphics[width=0.49\hsize]{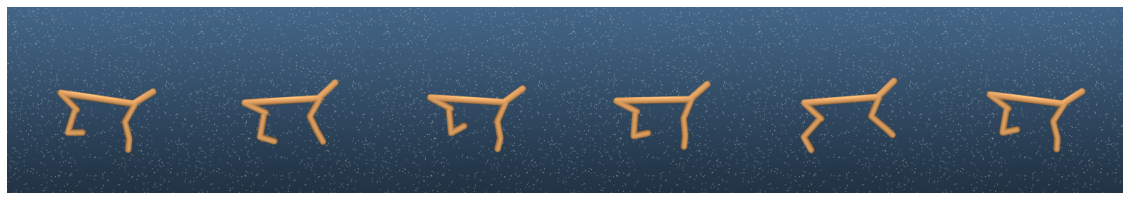}\hspace{-2.2mm}
    \includegraphics[width=0.49\hsize]{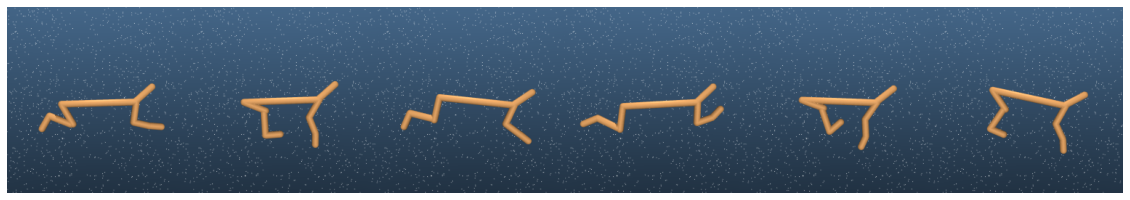}
    \includegraphics[width=0.49\hsize]{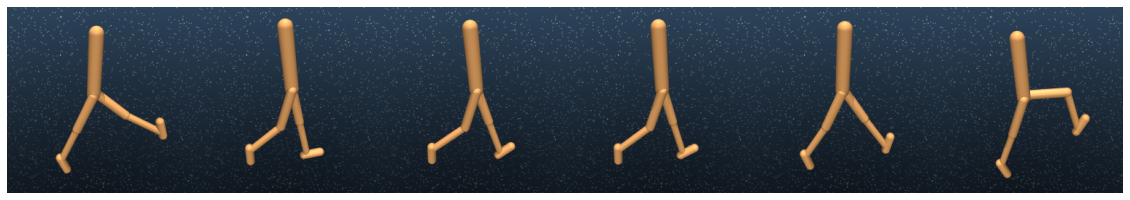}\hspace{-2.2mm}
    \includegraphics[width=0.49\hsize]{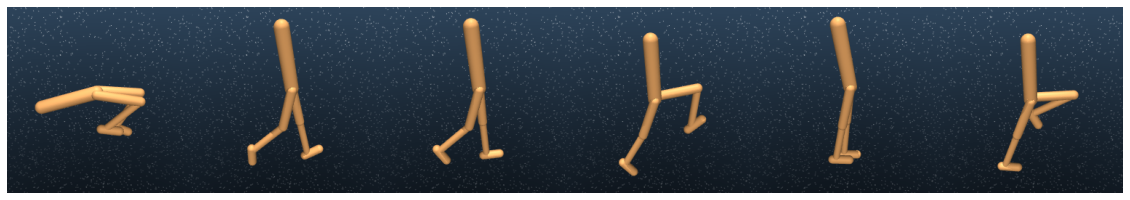}
    \includegraphics[width=0.49\hsize]{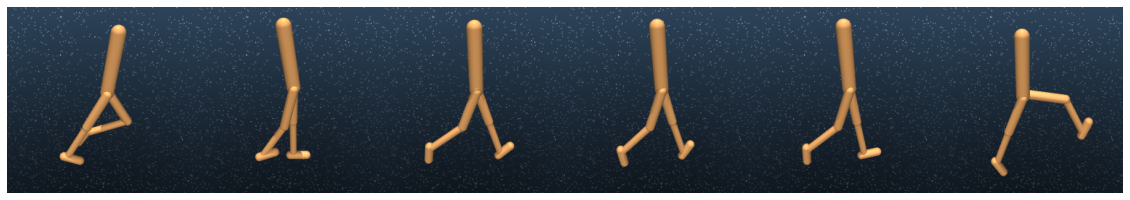}\hspace{-2.2mm}
    \includegraphics[width=0.49\hsize]{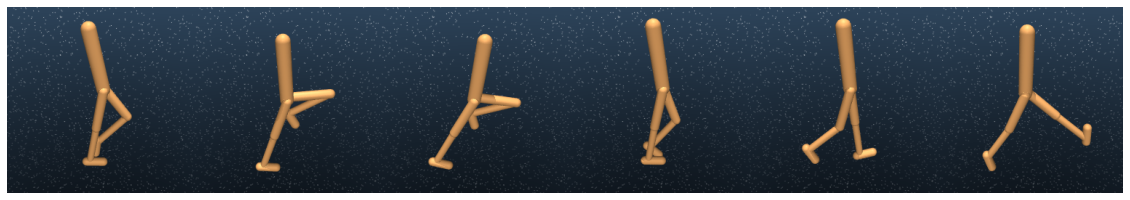}
    \includegraphics[width=0.49\hsize]{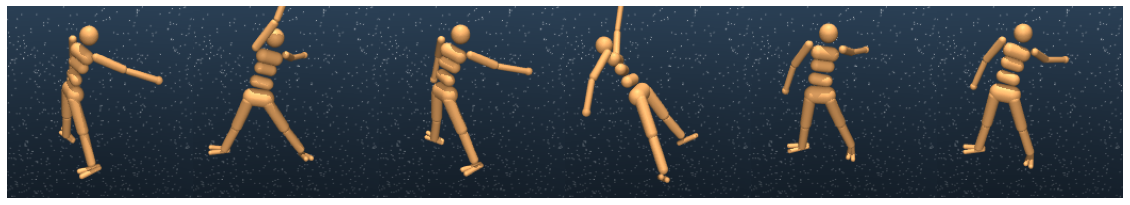}\hspace{-2.2mm}
    \includegraphics[width=0.49\hsize]{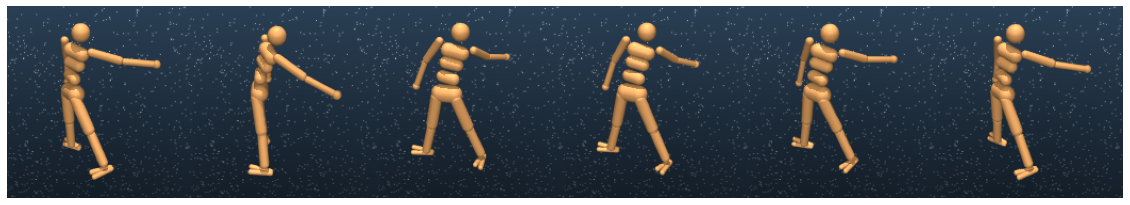}
    \includegraphics[width=0.49\hsize]{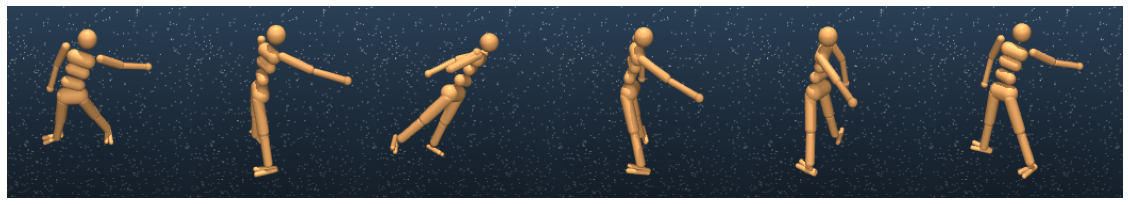}\hspace{-2.2mm}
    \includegraphics[width=0.49\hsize]{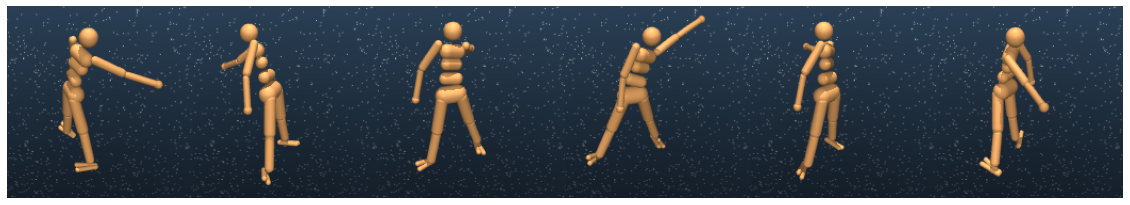}
    \caption{Uncurated samples computed with the trained CNF on product manifolds representing robot's state space: Cheetah (top), Walker (middle), and Humanoid (bottom). }
    \label{fig:robots_samples_apdx}
\end{figure}

\begin{figure}
    \centering
    \includegraphics[width=0.49\hsize]{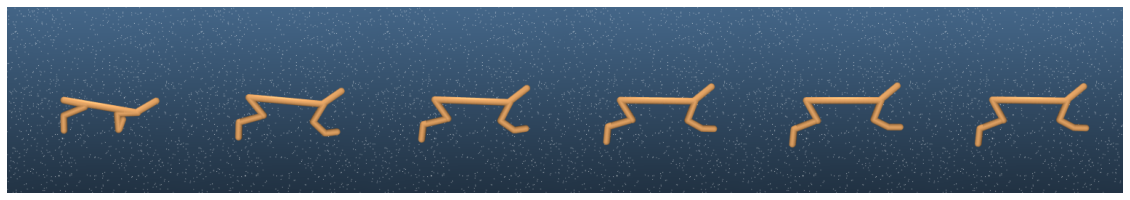}
    \includegraphics[width=0.49\hsize]{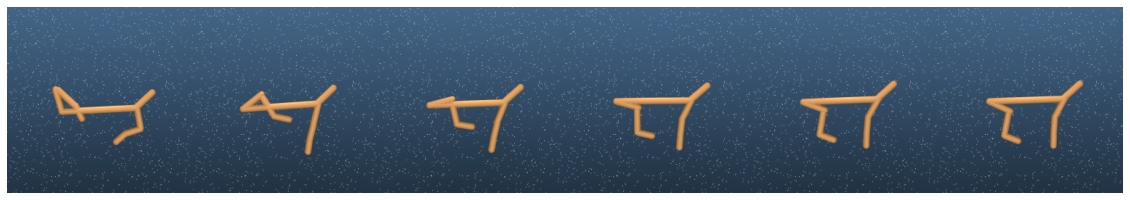}
    \includegraphics[width=0.49\hsize]{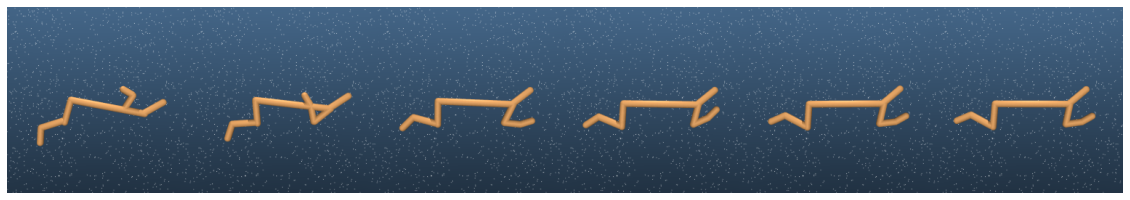}
    \includegraphics[width=0.49\hsize]{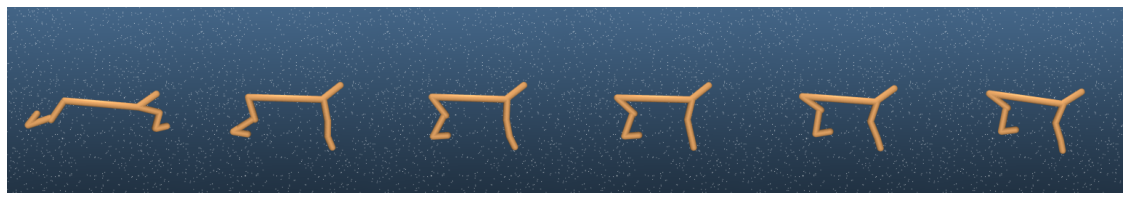}
    \includegraphics[width=0.49\hsize]{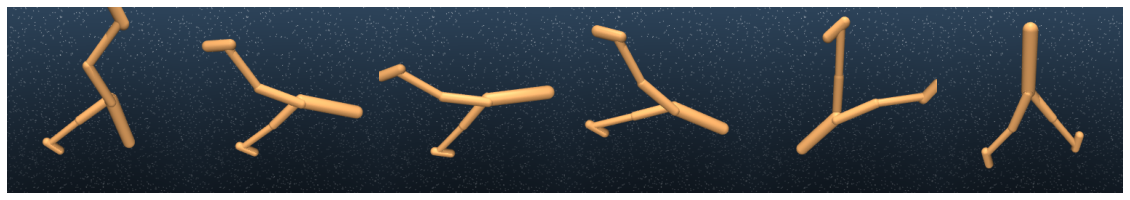}
    \includegraphics[width=0.49\hsize]{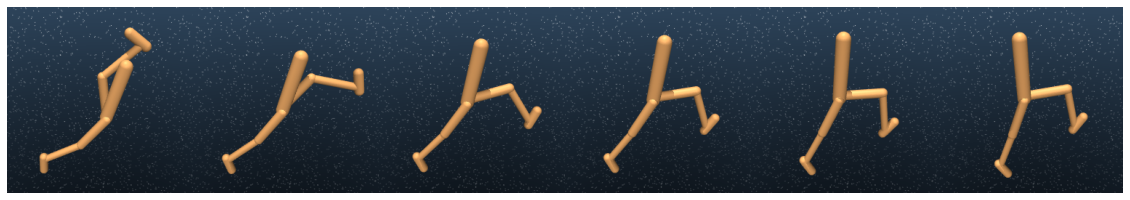}
    \includegraphics[width=0.49\hsize]{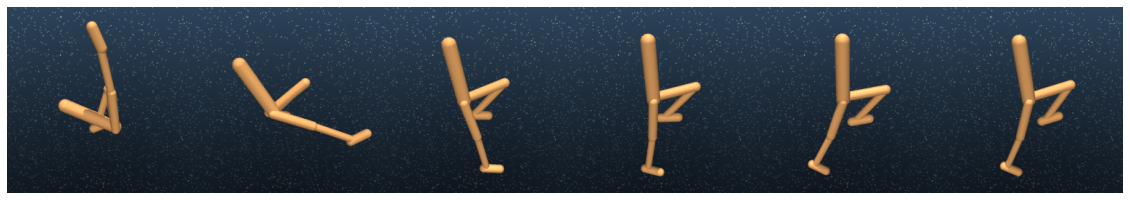}
    \includegraphics[width=0.49\hsize]{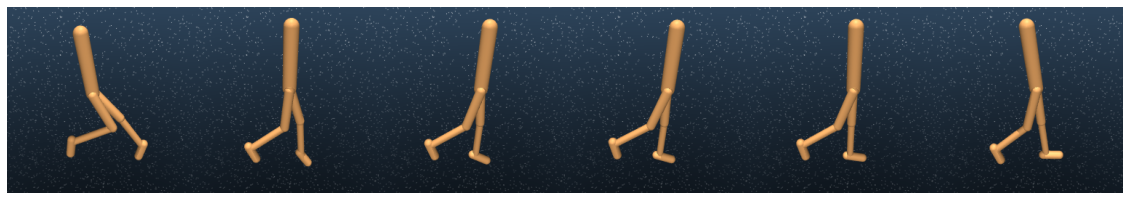}
    \includegraphics[width=0.49\hsize]{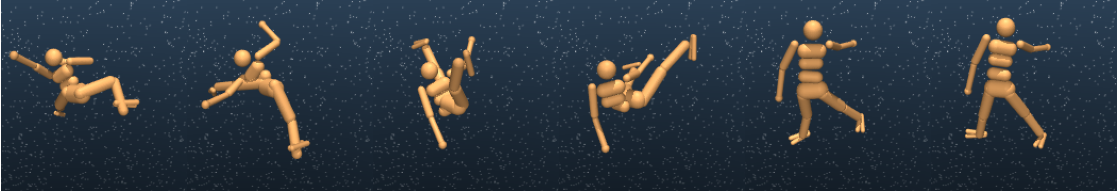}
    \includegraphics[width=0.49\hsize]{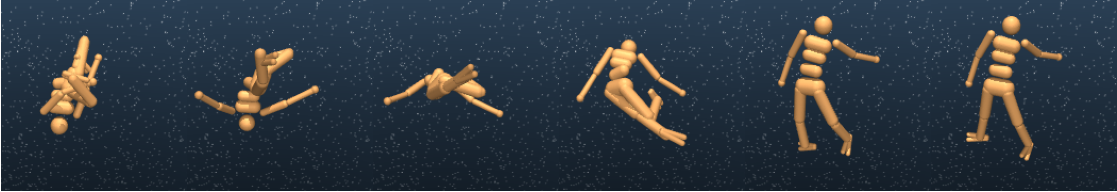}
    \includegraphics[width=0.49\hsize]{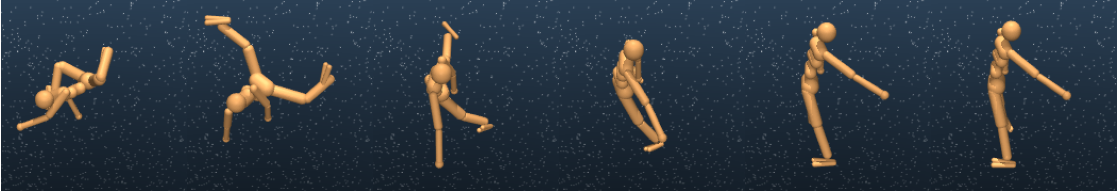}
    \includegraphics[width=0.49\hsize]{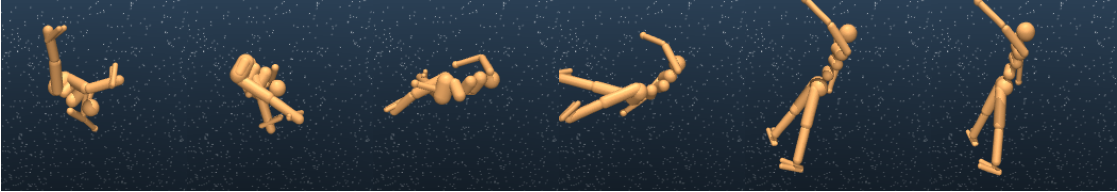}
    \caption{Noise to data paths computed with the trained CNF on product manifolds representing robot's state space:Cheetah (top), Walker (middle), and Humanoid (bottom).}
    \label{fig:robots_noise_to_data_apdx}
\end{figure}

We provide more uncurated samples and interpolations of Cheetah, Walker and Humanoid poses in \cref{fig:robots_samples_apdx} and \cref{fig:robots_noise_to_data_apdx}.

\end{document}